\documentclass[11pt]{article}

\usepackage[final]{acl}
\usepackage{times}
\usepackage{latexsym}

\usepackage[T1]{fontenc}
\usepackage[utf8]{inputenc}
\usepackage{microtype}
\usepackage{inconsolata}
\usepackage{graphicx}

\usepackage{rotating}   
\usepackage{booktabs}   
\usepackage{tabularx}   
\usepackage{makecell}   
\usepackage{amsmath}    
\usepackage{amssymb}    
\usepackage{bm}         

\usepackage{amssymb}

\usepackage{times}
\usepackage{soul}
\usepackage{url}
\usepackage{amsmath}
\usepackage{amsthm}
\usepackage{booktabs}
\usepackage{algorithm}
\usepackage{algorithmic}
\usepackage[switch]{lineno}
\usepackage{booktabs}
\usepackage{fontawesome}
\usepackage{tikz}
\usepackage{inconsolata}
\usepackage{pifont}
\usepackage[edges]{forest}
\usepackage{makecell}
\usepackage{cleveref}
\usepackage{setspace}
\definecolor{hidden-draw}{RGB}{205, 44, 36}
\definecolor{hidden-blue}{RGB}{194,232,247}
\definecolor{hidden-orange}{RGB}{243,202,120}
\definecolor{hidden-yellow}{RGB}{255,229,204}
\definecolor{hidden-red}{RGB}{255,204,204}
\definecolor{hidden-draw}{RGB}{20,68,106}
\definecolor{hidden-pink}{RGB}{255,245,247}
\usepackage{multirow}

\title{
Supervised Fine-Tuning for Reasoning Large Language Models: A Survey}

\author{
\textbf{Yongcan Yu\textsuperscript{1,2 *}},
\textbf{Kuangpu Guo\textsuperscript{1,3 *}},
\textbf{Jian Liang\textsuperscript{1,2 $\dagger$}},\\
\textbf{Lingxiao He\textsuperscript{4}},
\textbf{Meng Wang\textsuperscript{4}},
\textbf{Ran He\textsuperscript{1,2}}
\\[0.1cm]
\textsuperscript{1} NLPR \& MAIS, Institute of Automation, Chinese Academy of Sciences\\
\textsuperscript{2} School of Artificial Intelligence, University of Chinese Academy of Sciences\\
\textsuperscript{3} University of Science and Technology of China
\textsuperscript{4} Meituan Inc.
\\[0.1cm]
\faGithub\;\ttfamily\normalsize\href{https://github.com/yuyongcan/Awesome-SFT-for-LRMs}{yuyongcan/Awesome-SFT-for-LRMs}\ \{yuyongcan0223,liangjian92\}@gmail.com\\
}

\begin{document}
\maketitle

\begingroup
\renewcommand{\thefootnote}{\fnsymbol{footnote}}
\footnotetext[1]{These authors contributed equally to this work.}
\footnotetext[2]{Corresponding author.}
\endgroup

\begin{abstract}
Large reasoning models have recently demonstrated remarkable capabilities in solving complex tasks, where supervised fine-tuning (SFT) on long chain-of-thought data serves as a crucial foundation for eliciting and enhancing their reasoning abilities.
Despite rapid progress in both improving and analyzing reasoning-oriented SFT, the field still lacks a systematic survey that consolidates its fast-growing literature.
To fill this gap, we present a comprehensive review of recent advancements in reasoning SFT, examining the literature through the dual lenses of methodological design and analytical investigation.
First, we review methodological improvements across the SFT pipeline and categorize them into data-centric approaches and algorithm-centric innovations.
Second, we reorganize analytical studies along three dimensions: data characteristics, optimization dynamics, and mechanistic insights.
Finally, by synthesizing current research focuses and remaining bottlenecks, we outline promising future directions for reasoning SFT.
We hope this survey deepens the understanding of reasoning SFT and paves the way for advanced reasoning models.
\end{abstract}

\section{Introduction}
Reasoning has become a defining capability of modern large language models (LLMs), enabling them to solve complex tasks that require multi-step problem decomposition, intermediate verification, and logical abstraction~\citep{kahneman2011thinking,huang2023towards,sun2025survey,wang2026mitigating,wang2025comprehensive}.
Such capabilities are central to a wide range of domains, including mathematical problem solving~\citep{ahn2024large,wang2026survey}, code generation~\citep{zheng2023survey,jiang2026survey,lu2026worldcoder}, agents~\citep {xu2026train,lu2025deepresearch,xu2026dual}, question-answering~\citep{xue2024question,yu2025test}, and robotics~\citep{ma2026survey,sapkota2025vision}.
Early efforts primarily relied on Chain-of-Thought (CoT) prompting~\citep{wei2022chain,wang2024chain,chia2023contrastive,kojima2022large} to elicit reasoning behaviors from LLMs.
Although CoT prompting provides an effective interface for reasoning, it operates only at inference time and suffers from instability~\citep{wang2023towards,turpin2023language}.

To overcome these limitations, the paradigm has naturally transitioned toward Supervised Fine-Tuning (SFT)~\citep{lai2025survey,shen2024rethinking} on CoT trajectories to internalize reasoning capabilities within the model's parameters.
Early efforts in reasoning SFT predominantly focused on the dimension of data construction, exploring how to curate high-quality step-by-step rationales, typically through knowledge distillation from frontier models or automated synthesis pipelines~\citep{roziere2023code,li2024numinamath,xu2025llava,luo2025wizardmath}.
While such efforts provide an essential foundation, standard SFT based on vanilla behavior cloning also exhibits intrinsic limitations, including exposure bias~\citep{gan2026gft}, severe overfitting~\citep{huan2025does}, and limited generalization~\citep{Wu2026dft}.

To address these fundamental bottlenecks, the research community has recently pivoted toward algorithmic innovations that refine the optimization dynamics of SFT across diverse granularities, encompassing token- and sample-level signal modulation~\citep{kim2025sft, sanyal2025upweighting}, distribution constraints~\citep{zhu2025anchored}, and structural integration with reinforcement learning~\citep{jiang2026supervised}.
In parallel, a growing body of work has begun to analyze the behavior and mechanisms of reasoning SFT with long CoT data.
These studies investigate diverse factors such as data properties~\citep{ye2025limo}, optimization dynamics~\citep{ren2026rethinking}, and mechanistic interpretability~\citep{wu2025knowledge}, offering insights into when and why SFT improves reasoning.

Despite the recent surge of literature focusing on reasoning SFT, a systematic survey consolidating this fast-evolving landscape is still conspicuously absent.
While several existing reviews attempt to summarize post-training paradigms for reasoning models, they predominantly emphasize RL-based frameworks~\citep{zhou2025reinforced,sun2025reinforcement,xu2025toward} or restrict their examination of SFT to its alignment or synergy with RL pipelines~\citep{jiang2026supervised}.
Another stream of literature addresses reasoning-oriented SFT, yet focuses almost exclusively on early-stage data construction~\citep{chen2025towards,wang2026survey,li2025system}.
Consequently, a holistic consolidation of recent non-trivial algorithmic innovations and analytical insights remains lacking.

To fill this gap, this survey presents a systematic and timely overview of contemporary research, encompassing both methodological designs and analytical insights.
On the methodological front, we establish a structured taxonomy spanning data-centric improvements and algorithm-centric innovations.
On the analytical front, we synthesize and reorganize highly fragmented empirical findings into three core dimensions: data characteristics, optimization dynamics, and mechanistic analysis.
Furthermore, we identify open challenges and chart promising future directions to guide subsequent explorations in this domain.
Ultimately, we hope this work serves as a foundational roadmap to navigate and inspire future advancements in reasoning SFT.

\tikzstyle{my-box}=[
    rectangle,
    draw=hidden-draw,
    rounded corners,
    text opacity=1,
    minimum height=1.5em,
    minimum width=5em,
    inner sep=2pt,
    align=center,
    fill opacity=.5,
    line width=0.8pt,
]
\tikzstyle{leaf}=[my-box, minimum height=1.5em,
    text=black, align=left,font=\normalsize,
    inner xsep=2pt,
    inner ysep=4pt,
    line width=0.8pt,
]
\begin{figure*}[h!]
    \centering
    \resizebox{\textwidth}{!}{
        \begin{forest}
            forked edges,
            for tree={
                grow=east,
                reversed=true,
                anchor=base west,
                parent anchor=east,
                child anchor=west,
                base=left,
                font=\large,
                rectangle,
                draw=hidden-draw,
                rounded corners,
                align=left,
                minimum width=4em,
                edge+={darkgray, line width=1pt},
                s sep=7pt,
                inner xsep=2pt,
                inner ysep=3pt,
                line width=0.8pt,
                ver/.style={rotate=90, child anchor=north, parent anchor=south, anchor=center},
            },
            where level=1{text width=6em,font=\normalsize}{},
            where level=2{text width=9em,font=\normalsize}{},
            where level=3{text width=8em,font=\normalsize}{},
            where level=4{text width=10em,font=\normalsize}{},
            [
                {Supervised Fine-tuning for Reasoning Large Language Models},fill=hidden-yellow!70,ver
                [
                    {Methods (\S \ref{sec:methods})},fill=hidden-red!70
                    [
                        {Data-Centric \\ Methods (\S \ref{sec:methods_data})},fill=hidden-blue!70
                        [
                            {Data Augmentation},fill=hidden-blue!70
                            [
                                {\cite{lin2025debunk}, \cite{gan2026gft}, \cite{chen2026nft}, \cite{liu2026rift}}, leaf, text width=42em,fill=hidden-blue!70
                            ]   
                        ]
                        [
                            {Data Refinement},fill=hidden-blue!70
                            [
                                {\cite{gupta2025selective}, \cite{zhao2026mind}, \cite{wang2026patch}, \cite{lin2026surgical}}, leaf, text width=42em,fill=hidden-blue!70
                            ] 
                        ]
                    ]
                    [
                        {Algorithm-Centric \\ Methods  (\S \ref{sec:methods_Objective})},fill=hidden-blue!70
                        [
                            {Token-Level \\ Optimization}, fill=hidden-blue!70
                            [
                                {Reweighting: \cite{kim2025sft}, \cite{gong2025vcore}, \cite{Wu2026dft}, \cite{diao2026entropy}, \\  \cite{wang2026gradients},\cite{yu2026probability}, \cite{lin2025sft}, \cite{li2026getting}, 
                                \cite{zhao2026gift}, \\ 
                                \cite{zhang2026good}. \\
                                Selection: \cite{pang2025token}, \cite{qin2025sstoken}, \cite{li2026token}, \cite{ruan2025enhancing}, \\ 
                                \cite{yang2026explainable}, \cite{ma2025ts}, \cite{liu2026profit}, \cite{chen2026sed}, \cite{shi2025rethinking}}, leaf, text width=42em,fill=hidden-blue!70
                            ]
                        ]
                        [
                            {Sample-Level \\ Optimization}, fill=hidden-blue!70
                            [
                                 {Reweighting: \cite{sanyal2025upweighting}, \cite{tang2025aligning}, \cite{qin2025supervised}, \\
                                 \cite{shu2026reference}, \cite{ming2025one}, \cite{zhang2025holdout}, \cite{pan2025scalebio}. \\
                                 Selection: \cite{liu2024selectit}, \cite{zhang2026best}, \cite{yang2026reasoning},\cite{lu2026does}, \\ 
                                 \cite{xia2024less}, \cite{wan2026computational}, \cite{fu2026t}, \cite{wang2026step}, \cite{wang2025winning}}, leaf, text width=42em,fill=hidden-blue!70
                            ]
                        ]        
                        [
                            {Distribution-Level \\ Regularization}, fill=hidden-blue!70
                            [
                                {\cite{zhu2025anchored}, \cite{zhu2025proximal}, \cite{wang2026stabilizing}, \cite{li2025preserving}, \cite{wang2026learning}}, leaf, text width=42em,fill=hidden-blue!70
                            ]
                        ]
                        [
                            {Integration with RL}, fill=hidden-blue!70
                            [
                                {\cite{yang2026learning}, \cite{lv2025towards}, \cite{he2025amft}, \cite{zhang2026onpolicy}, \cite{wang2026learning}, \\
                                \cite{chen2025beyond},  \cite{guan2025recall}, \cite{fu2026srft}, \cite{ma2025learning}, \cite{zhu2026bridging}, \\
                                \cite{liu2025empowering}, \cite{su2025trust}}, leaf, text width=42em,fill=hidden-blue!70
                            ]
                        ]
                    ]
                ]
                [
                    {Analyses (\S \ref{sec:study})},fill=hidden-red!70
                    [
                        {Data-Centric \\Analysis  (\S \ref{sec:std_data})},fill=hidden-blue!70
                        [
                            {\cite{yang2025emperor}, \cite{yao2026compositional}, \cite{lin2025debunk}, \cite{yang2025demystifying}, \cite{shen2025long},
                            \cite{wu2026when}, \\
                            \cite{zhang2026why},
                            \cite{chandra2025shape},
                            \cite{yang2026reasoning},
                            \cite{zhao2025more},
                            \cite{luo2025deconstructing}, 
                            \cite{ye2025limo}, \\
                            \cite{wu2025beyond},
                            \cite{muennighoff2025s1},
                            \cite{li2025language},
                            \cite{li2026role}, 
                            \cite{chen2025sft}, 
                            \cite{do2025defines},\\
                            \cite{yu2025reassessing},
                            \cite{huang2026fine},
                            \cite{xue2026supervised}
                            }, leaf, text width=51.5em,fill=hidden-blue!70
                        ]
                    ]
                    [
                        {Optimization-Centric \\Analysis  (\S \ref{sec:std_opt})},fill=hidden-blue!70
                        [
                            {\cite{kopiczko2026data}, \cite{ren2026rethinking}, \cite{jin2025rl}, \cite{ding2025rethinking}, \cite{li2026getting},
                            \cite{lin2025sft}, \\
                            \cite{zhang2026bread},
                            \cite{li2025small},
                            \cite{ding2025micota},
                            \cite{yu2025reassessing}
                            }, leaf, text width=51.5em,fill=hidden-blue!70
                        ]
                    ]
                    [
                        {Mechanistic \\ Analysis  (\S \ref{sec:std_mec})},fill=hidden-blue!70
                        [
                            {\cite{yang2025emperor}, \cite{wu2025knowledge}, \cite{yue2026does}, \cite{matsutani2026rl},
                            \cite{yao2026compositional},
                            \cite{chu2025sft}, \\
                            \cite{huan2025does},
                            \cite{jin2025rl},
                            \cite{lai2025reinforcement},
                            \cite{chen2025retaining},
                            \cite{yang2025demystifying},
                            \cite{ren2026rethinking}, \\
                            \cite{tian2026learning},
                            \cite{lin2025debunk},
                            \cite{yue2026does},
                            \cite{matsutani2026rl},
                            \cite{zhang2025interplay}, 
                            \cite{chen2025sft}, \\
                            \cite{chen2026the},
                            \cite{yu2025reassessing},
                            \cite{limozin2026sft},
                            \cite{ding2025rethinking},
                            \cite{li2026getting},
                            \cite{kang2026quagmires}
                            }, leaf, text width=51.5em,fill=hidden-blue!70
                        ]
                    ]
                ]
            ]
        \end{forest}
    }
    \caption{Taxonomy of methodological and analytical research in SFT for reasoning LLMs.}
    \label{fig:survey}
\end{figure*}
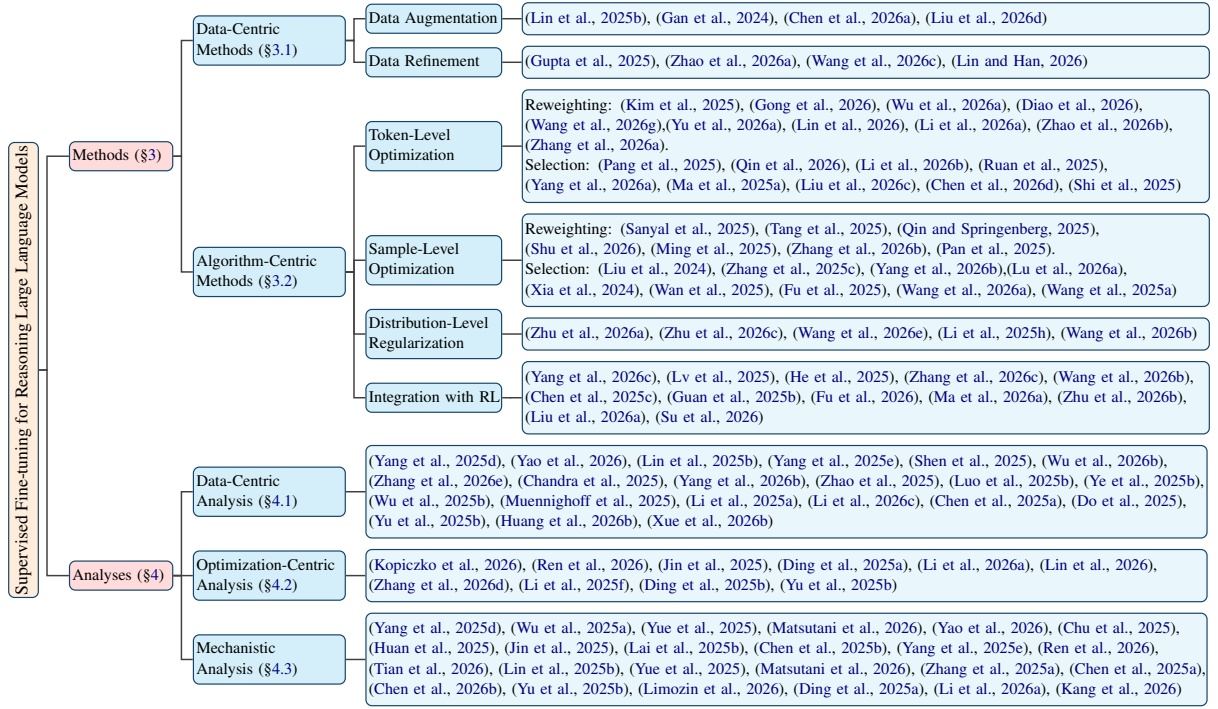
\section{Preliminary}
SFT serves as the foundational paradigm for aligning LLMs with human intent using expert demonstrations.
Formally, given a language model parameterized by $\theta$ and a training dataset $\mathcal{D} = \{(x^{(i)}, y^{(i)})\}_{i=1}^N$, standard SFT minimizes the Negative Log-Likelihood (NLL) objective:
\begin{equation*}
    \mathcal{L}_{\text{SFT}}(\theta) = - \mathbb{E}_{(x,y) \sim \mathcal{D}} \left[ \sum_{t=1}^{|y|} \log \pi_\theta(y_t | y_{<t}, x) \right],
\end{equation*}
where $x$ denotes the input prompt, $y$ represents the target response sequence, and $\pi_\theta$ specifies the model policy. 
Within the scope of this survey, we center our focus exclusively on \textit{reasoning SFT}.
Distinct from standard instruction-following, the target response $y$ is structurally decomposed into an extensive CoT rationale $r$ followed by the final answer $a$, \textit{i.e.,} $y=[r;a]$.
Optimizing $\mathcal{L}_{\text{SFT}}$ over protracted reasoning paths compels the model to internalize intrinsic causal structures, thereby capturing genuine problem-solving mechanics rather than exploiting surface-level correlations.
\section{Reasoning SFT: Methods}
\label{sec:methods}

\subsection{Data-Centric Methods}
\label{sec:methods_data}

\paragraph{Data Augmentation.}
To actively ensure robust internalization and prevent the model from overfitting to specific reasoning formats, data augmentation~\citep{maharana2022review} has emerged as a critical data-centric strategy for enriching the diversity of training distributions.
At the input level, \cite{lin2025debunk} leverage systematic prompt diversification to break the model's anchoring to surface-level linguistics.
For target responses, GFT~\cite{gan2026gft} mitigates single-path dependency by pairing queries with a hybrid ensemble of expert, teacher, and self-generated trajectories.
Furthermore, recent methodologies have shifted from purely positive imitation toward extracting dense contrastive signals from negative samples.
For instance,  NFT~\cite{chen2026nft} constructs an implicit negative policy by jointly modeling both successful and failed reasoning responses, utilizing the contrast to actively push the policy away from common hallucination patterns and logical fallacies.
Taking a reward-calibrated approach, RIFT~\cite{liu2026rift} assigns a constant positive reward to correct reasoning traces, while explicitly allocating a constant negative reward to incorrect trajectories.

\paragraph{Data Refinement.}
Despite the benefits of augmented diversity, directly training on off-policy expert trajectories often introduces a distributional mismatch that can degrade a model's pre-existing capabilities. 
To address this, data refinement aims to seamlessly align the training corpus with the LLM's intrinsic distribution~\cite{gupta2025selective, zhao2026mind, wang2026patch, lin2026surgical}.    
Recent self-alignment frameworks~\cite{gupta2025selective, zhao2026mind} substitute standard expert demonstrations with a combination of the model's inherently correct generations and its successfully rewritten correct responses.
Specifically, the rewriting process leverages reference answers to guide the model in reconstructing valid, in-distribution reasoning trajectories.
Alternatively, SPOT~\cite{lin2026surgical} employs a stronger teacher model for minimal-edit step corrections and utilizes a decoupled objective to reinforce rectified reasoning steps and penalize logical errors.
To overcome the diversity collapse inherent in heuristic templates, another work~\cite{wang2026patch} formulates data refinement as a policy learning task, optimizing semantic diversity and alignment via reinforcement learning under a task-consistency gate.

\subsection{Algorithm-Centric Methods}
\label{sec:methods_Objective}
Alongside data-centric strategies, algorithm-centric methodologies offer a complementary axis for enhancing standard SFT by directly modifying the underlying optimization dynamics. To ensure a mutually exclusive and clean taxonomy, we adopt \textit{optimization granularity}---progressing from token-level and sample-level modulation to distribution-level regularization and RL integration---as our primary split criterion. While alternative perspectives could categorize these methods based on their underlying data sources or heuristics (e.g., model confidence, sample difficulty, semantic rules, or reference policy divergence), structuring by granularity provides a unified mathematical lens. To bridge these perspectives and clarify potential conceptual overlaps across heuristic motivations, we provide a comprehensive 2D map as shown in Table~\ref{tab:2d_map} in the supplementary material, illustrating the intersection of optimization granularity and data heuristics. Accordingly, this subsection systematically reviews these algorithmic advancements along the granularity axis.

Formally, fine-grained signal modulation can be unified under a generalized optimization objective:
\begin{equation*}
\begin{split}
\mathcal{L}_{\text{SFT}}^g(\theta) =& - \frac{1}{|\mathcal{D}|} \sum_{i=1}^{|\mathcal{D}|}  \color{red}{\bm{\alpha}_i} \cdot \\
&\left[ \sum_{t=1}^{|y^{(i)}|} {\color{blue}{\bm{w}_{t}^{(i)}}} \log \pi_\theta(y_t^{(i)} | y_{<t}^{(i)}, x^{(i)}) \right], 
 \end{split}
\end{equation*}
where $\color{blue}{w_{t}^{(i)}}$ and $\color{red}{\alpha_i}$ denote the dynamic token-level and sample-level weights, respectively.
By actively calibrating these coefficients alongside global regularization and RL integration, recent frameworks systematically mitigate the empirical limitations of standard supervised alignment.

\paragraph{Token-Level Optimization.}
We first review token-level optimization, where the learning dynamics are modulated by the token weight $\color{blue}{w_{t}^{(i)}}$ to prioritize suitable reasoning tokens and mitigate overfitting. 
Within the reweighting paradigm, recent works dynamically scale token updates based on gradient consistency between the individual token and the global batch~\cite{gong2025vcore}, the predicted probabilities~\cite{Wu2026dft}, or the output distribution entropy~\cite{diao2026entropy}. 
However, these methods expose a critical limitation: relying solely on probability or loss struggles to distinguish inherently hard reasoning steps from mere noisy tokens. 
To address the ambiguity of low-probability tokens, DEFT~\cite{wang2026gradients} introduces an entropy-driven trust gate that transitions the optimization focus from exploring uncertain regions to sharpening confident predictions. 
Building on a similar motivation, RANKTUNER~\cite{yu2026probability} aggregates probability and entropy into a unified ranking metric, systematically upweighting critical low-probability, low-entropy targets. 
From an optimization perspective, TALR~\cite{lin2025sft} applies a dynamic, normalized weight to each token. 
Specifically, the token weight is formulated as $w_i = \frac{\exp(-\ell_i/\tau)}{\sum_j \exp(-\ell_j/\tau)}$, where $\ell_i$ is the token-level loss and $\tau$ represents the median batch loss during training.

A parallel line of research tailors the SFT objective via token reweighting to establish an optimal initialization policy for subsequent RL alignment.
These methods recognize a fundamental trade-off: while aggressively upweighting specific tokens maximizes SFT imitation, it risks distributional collapse that destroys the exploration potential crucial for the RL cold start. 
To mitigate this collapse and preserve exploration potential, recent frameworks maintain the pre-trained probability landscape by amplifying gradients on uncertain tokens via sequence-relative weighting~\cite{li2026getting}. 
Alternatively, another approach~\cite{zhao2026gift} upweights correct tokens via a finite-temperature Gibbs distribution, which explicitly preserves the relative probability mass of alternative tokens to mitigate distributional collapse.
This highlights a crucial shift: from sharp, loss-driven imitation for standalone SFT, to soft, distribution-preserving scaling for RL initialization.
To further bridge the gap between SFT and RL, PEAR~\cite{zhang2026good} employs multi-level importance sampling, discounted suffix weighting, and negative sample suppression to construct a superior initialization for online RL.

Within the discrete token selection paradigm, recent methods extract optimal learning tokens through diverse filtering criteria. 
Some works explicitly retain tokens exhibiting the highest training loss~\cite{kim2025sft} or the largest parameter update norms~\cite{ma2025ts}. 
A parallel line of work identifies high-quality tokens via cross-model loss discrepancies, comparing superior and inferior models to preserve critical tokens~\cite{pang2025token}, or contrasting safety-degraded and utility-oriented policies to eliminate vulnerabilities~\cite{li2026token}. 
Alternative frameworks restrict gradient updates exclusively to critical tokens identified through semantic-aware response-to-prompt attention~\cite{qin2025sstoken}, counterfactual perturbations~\cite{ruan2025enhancing}, or multi-dimensional explainability metrics~\cite{yang2026explainable}.

Beyond gradient and semantic signals, further approaches define selective optimization strategies based on the model's predictive confidence, policy-driven sampling, or predefined trajectory structures.
ProFit~\cite{liu2026profit} consolidates established knowledge by computing the loss solely on high-probability tokens, whereas SED-SFT~\cite{chen2026sed} encourages generative diversity by selectively optimizing low-confidence tokens associated with multiple plausible branches. 
To explicitly incorporate on-policy exploration into the SFT phase, OTR~\cite{ming2025one} formulates token generation as a single-step reinforcement learning trajectory, dynamically rewarding ground-truth tokens while penalizing incorrect samples drawn from the current policy distribution. 
Finally, from a structural standpoint, SFTKey~\cite{shi2025rethinking} completely masks the intermediate CoT rationale to prevent verbose reasoning trajectories from overshadowing core conclusions, thereby explicitly restricting optimization solely to the answer tokens.

\paragraph{Sample-Level Optimization.}
Transitioning to sample-level optimization, the learning dynamics are governed exclusively by the dynamic sample weight $\color{red}{\alpha_i}$. 
Within the continuous reweighting framework, one prominent branch calibrates the scalar $\color{red}{\alpha_i}$ based on the intrinsic difficulty of the sample. 
For example, FLOW~\cite{sanyal2025upweighting} upweights ``easy'' samples with low pre-training losses to mitigate catastrophic forgetting, whereas BFT~\cite{tang2025aligning} prioritizes ``hard'' sequences featuring continuous low-probability spans to force the absorption of novel knowledge.

Another direction establishes a theoretical equivalence between offline SFT and online RL by reweighting static samples to simulate the dynamic distribution shifts inherent to RL.
This theoretical equivalence to RL is realized by deriving adaptive importance weights from policy trajectory ratios~\cite{qin2025supervised} or scaling prompt-normalized exponential rewards~\cite{shu2026reference}.
Moving beyond predefined mathematical formulations, a final line of research automates sample utility evaluation and data composition.
These frameworks dynamically optimize instance weights based on how much their loss drops when shown high-quality reference examples as context~\cite{zhang2025holdout}, or by framing sample weighting as a minimax bilevel optimization problem to automatically learn optimal multi-source data proportions~\cite{pan2025scalebio}.

Complementary to continuous modulation, discrete sample selection employs hard pruning to isolate optimal training instances.
To align the training corpus with the model's inherent distribution, recent frameworks evaluate internal uncertainty to identify high-quality instructions~\cite{liu2024selectit}, or select teacher-generated responses that maximize the student model's assigned probability~\cite{zhang2026best}.
Furthermore, to align with the model's current learning capacity, other methods~\cite{yang2026reasoning, lu2026does} explicitly filter out excessively challenging instances to focus optimization on moderate-difficulty zones.

Beyond heuristic difficulty assessments, a parallel paradigm formulates sample selection through rigorous mathematical optimization and fine-grained structural dynamics. 
To curate datasets under strict computational budgets, optimization-driven frameworks prioritize samples whose gradients closely align with predefined target examples~\cite{xia2024less} or by framing efficient data selection as a surrogate-penalized bilevel optimization problem~\cite{wan2026computational}. 
To mitigate structural biases, further studies refine sample scoring by prioritizing informative tokens that are robust to embedding noise~\cite{fu2026t}, or by causally disentangling step-length confounders to calibrate branching first-token probabilities~\cite{wang2026step}.
Finally, moving beyond isolated sample- or token-level filters, Q-Tuning~\cite{wang2025winning} unifies both granularities via an error-uncertainty diagnostic framework, dynamically discarding entirely noisy samples while applying token pruning exclusively to partially misunderstood instances.

\paragraph{Distribution-Level Regularization.}
Stepping beyond localized token and sample-level interventions, distribution-level regularization constrains the global policy distribution to mitigate catastrophic forgetting and preserve pre-trained capabilities. 
To restrict deviations from the reference policy, recent methodologies explicitly bound the optimization trajectory by incorporating a standard Kullback-Leibler (KL)~\citep{kullback1951kullback} divergence penalty~\cite{zhu2025anchored}, applying hard gradient clipping beyond predefined divergence thresholds~\citep{zhu2025proximal}, or enforcing a strict upper bound on iterative KL divergence through dynamic moving anchors~\citep{wang2026stabilizing}.
To prevent diversity degradation and maintain exploration potential, a parallel line of research leverages entropy-aware modulation.
These frameworks sustain probability mass across diverse reasoning trajectories by either simulating reverse KL minimization via a sparse meta-controller~\cite{li2025preserving}, or applying asymmetric regularization that explicitly targets high-entropy reasoning steps while strictly imitating low-entropy factual knowledge~\cite{wang2026learning}.

\paragraph{Integration with RL.}
Unifying SFT and RL avoids the imitation bottleneck of pure SFT and the optimization instability of isolated RL, enabling models to expand their capability boundaries while refining reasoning trajectories.
To ensure a clear structural boundary, this subsection is strictly confined to approaches utilizing an explicit SFT loss ($\mathcal{L}_{\text{SFT}}$) alongside RL objectives. 
Frameworks that use demonstrations merely as exploration anchors are discussed in Appendix~\ref{Integration with RL add}.

One prominent paradigm dynamically harmonizes these objectives via continuous loss weighting. 
For instance, some works adaptively calibrate SFT and RL coefficients based on real-time metrics such as response correctness~\cite{yang2026learning, lv2025towards} or diversity~\cite{yang2026learning}.
AMFT~\cite{he2025amft} employs a meta-gradient weight controller to dynamically learn the optimal balance between supervised imitation and reinforcement exploration.
Incorporating predictive uncertainty as a guiding metric, RED~\cite{guan2025recall} dynamically modulates learning weights via entropy scheduling to balance exploration and path refinement. 
Building upon this uncertainty-aware perspective, SRFT~\cite{fu2026srft} integrates an entropy-regularized SFT objective with off-policy RL, stabilizing knowledge distillation without constraining the policy's exploratory potential.
Alternatively, CHORD~\cite{zhang2026onpolicy} employs deterministic schedules to smoothly transition from SFT to RL, and BRIDGE~\cite{chen2025beyond} enforces structural decoupling by restricting SFT updates to lightweight adapters.

Alternatively, discrete sample-wise routing allocates instances based on difficulty or correctness.
To systematically overcome performance plateaus, some methods route excessively hard~\cite{ma2025learning} or completely failed prompts~\cite{zhu2026bridging} exclusively to SFT or multi-teacher distillation.
Conversely, successfully resolved prompts trigger RL to consolidate reasoning pathways, occasionally incorporating visual refiners in DyME~\cite{liu2025empowering} or contrastive objectives in DYPO to stabilize exploration.
Advancing this integration to a finer granularity, TRAPO~\cite{su2025trust} applies supervised learning to adaptively selected expert prefixes, while employing standard RL to optimize the subsequent model-generated rollouts.

\section{Reasoning SFT: Analyses}
\label{sec:study}

\subsection{Data-Centric Analysis}
\label{sec:std_data}

\paragraph{The Impact of CoT Length.}
Characterized by its intuitive and quantifiable nature, CoT length serves as a natural entry point for investigating the empirical behavior of reasoning SFT.
Several studies have shown that CoT supervision is crucial for reasoning: models fine-tuned without CoT suffer substantial performance degradation or poor generalization \citep{yang2025emperor, yao2026compositional, zhang2025improve, lin2025debunk}.
A common view in the literature is that longer CoTs are more beneficial than shorter ones, yielding stronger performance and better OOD generalization \citep{yang2025demystifying}.
\citet{shen2025long} further shows that CoT length matters more than problem difficulty in determining model performance.
However, \citet{wu2026when} argue that CoT length follows an inverted-U relationship with model accuracy: both overly short and overly long CoTs are suboptimal, with the optimal length increasing for harder tasks but decreasing as model capability improves.


\paragraph{Distribution, Difficulty, and Diversity.}
Data distribution plays a critical role in reasoning SFT~\citep{zhang2026why,chandra2025shape,yang2025emperor,yang2026reasoning,zhao2025more,luo2025deconstructing}. 
\citet{chandra2025shape} show that LLM-paraphrased CoTs, even when containing errors, can outperform manually constructed data, suggesting that distributional alignment may matter more than surface-level correctness. 
This view is further supported by studies emphasizing teacher--student alignment in long-CoT distillation~\citep{yang2025emperor,luo2025deconstructing,wu2025beyond}.
To mitigate distribution mismatch, recent works further explore stronger or more adaptive data sources, including advanced reasoning models~\citep{zhao2025more}, RL-generated trajectories~\citep{zhang2026why}, and teacher-student collaborative data construction~\citep{huang2026fine}.

Beyond distributional alignment, problem difficulty and data diversity also shape SFT effectiveness. 
\citet{ye2025limo,muennighoff2025s1} favor harder and more diverse samples when constructing reasoning datasets, while \citet{wu2025beyond} advocate filtering out overly simple problems and maintaining response diversity through Levenshtein-distance filtering. 
In addition, a recent study highlights the critical role of prompt diversity in improving generalization~\citep{lin2025debunk}.

\paragraph{Reasoning Pattern.}
Beyond surface-level data properties, another line of work examines how the internal reasoning patterns of CoTs affect SFT, including their structural organization, exploration behaviors, and error-correction processes.
An early study shows that disrupting the structure of CoTs leads to larger performance degradation than corrupting their surface content, highlighting the central role of reasoning structure in CoT-based training~\citep{li2025language}.
However, more complex or exploratory reasoning patterns are not always beneficial.
\citet{li2026role} find that stronger teacher models do not necessarily yield better distillation, and that teachers producing convergent and deductive trajectories can be more effective than those generating divergent, branch-heavy explorations.
Similarly, SFT on CoTs with ``Aha moments'' may degrade performance, suggesting that explicitly imitating complex patterns is not always helpful~\citep{chen2025sft}.

Beyond global reasoning patterns, the quality of individual reasoning steps also matters.
\citet{do2025defines} show that final-answer correctness alone is insufficient for data filtering, and that selecting examples based on intermediate-step relevance and step-to-step coherence can further improve performance.
Moreover, removing similar or repetitive steps can reduce token costs and improve performance, whereas retaining erroneous steps may help models learn verification, backtracking, and error correction~\citep{luo2025deconstructing}.

\paragraph{Data Quantity.}
Although large-scale model training is often guided by the assumption that more data improves performance~\citep{li2025misfitting}, reasoning SFT appears to be less dependent on data quantity. 
Several studies show that small but carefully curated datasets can substantially improve or elicit reasoning capabilities, with fewer than 1K examples often being sufficient~\citep{ye2025limo,muennighoff2025s1} and even as few as 10 examples working in extreme settings~\citep{yang2025emperor}. 
In VLM post-training, SFT performance has also been found to be relatively insensitive to data quantity~\citep{yu2025reassessing}. 
Overall, these findings suggest that the structure and quality of reasoning data matter more than its scale~\citep{zhao2025more}, and increasing SFT data may even hurt performance in some cases~\citep{chen2025sft}.

\subsection{Optimization-Centric Analysis}
\label{sec:std_opt}

\paragraph{Training Dynamics.}
Beyond data construction, reasoning SFT is highly sensitive to training dynamics, including optimization duration, checkpoint selection, and learning-rate control. 
These factors are crucial for balancing reasoning improvement, generalization, and capability preservation.

A central question is how extensively SFT should be performed. 
For standalone SFT, several studies suggest that sufficient optimization is necessary to internalize reasoning patterns.
\citet{kopiczko2026data,liu2026less} show that multi-epoch training is important for reasoning SFT, as insufficient optimization can substantially degrade performance. 
\citet{ren2026rethinking} further reveal a non-monotonic phenomenon, where OOD performance first decreases and then recovers as training proceeds. 

However, when SFT serves as initialization for subsequent RL, excessive optimization may be undesirable. 
In contrast to the recovery pattern observed by \citet{ren2026rethinking}, \citet{jin2025rl} find that OOD performance first improves and then declines, possibly due to differences in the evaluated tasks. 
Consistent with this concern, \citet{ding2025rethinking} and \citet{li2026getting} show that early-stage SFT checkpoints can provide better initialization for downstream RL. 
In addition, \citet{lin2025sft} find that smaller learning rates better balance in-distribution and OOD performance.


\paragraph{Model Capability.}
Reasoning SFT is strongly capability-dependent: Several studies reveal that weak models may struggle to learn the complex patterns encoded in long CoTs~\citep{zhang2026bread,ren2026rethinking,li2025small,ding2025micota}. 
However, within the domain of VLM reasoning, \citet{yu2025reassessing} observe that SFT with short CoT disproportionately benefits weaker models, yielding more pronounced improvements.
To improve learnability for smaller models, existing work explores easier supervision signals, such as mixing short-CoT data or trajectories from smaller teachers~\citep{li2025small}, or fine-tuning a medium-sized model as an intermediate teacher to generate data for smaller models~\citep{ding2025micota}.


\subsection{Mechanistic Analysis}
\label{sec:std_mec}

\paragraph{What Does SFT Actually Teach?}
A central question in reasoning SFT is whether it genuinely teaches new reasoning capabilities or mainly induces models to imitate the surface forms and behavioral patterns of reasoning traces. 

Several studies suggest that SFT may primarily reshape model behavior rather than create new reasoning abilities. 
\citet{yang2025emperor} show that even 10 content-corrupted examples can substantially improve reasoning performance, indicating that SFT may mainly induce format imitation. 
Similarly, models trained on golden shortest paths fail to acquire backtracking capabilities, in contrast to those trained with RLVR~\citep{wei2026provable,wen2026reinforcement}. 
\citet{wu2025knowledge} further argue that SFT mainly injects task knowledge into the model and may even compromise its reasoning capability.

In contrast, another line of work argues that SFT can expand the model's capability boundary rather than merely reshape its output format. 
From the perspective of Pass@k, several studies suggest that SFT can increase the set of problems that the model is capable of solving~\citep{yue2026does,matsutani2026rl}. 
Moreover, \citet{yao2026compositional} show that CoT-trained models can compose learned simple skills to solve unseen or more complex problems, providing evidence for genuine reasoning.

\paragraph{Memorization or Generalization?}
A pivotal inquiry in reasoning SFT is whether the observed improvements stem from the memorization of specific trajectories or the acquisition of transferable reasoning patterns.
A critical view is that SFT tends to memorize rather than generalize: \citet{chu2025sft} observe a continuous decline in OOD performance on Points24~\citep{zhai2024fine} and navigation tasks~\citep{yang2024v}, leading to the conclusion that ``SFT memorizes, RL generalizes''.
Related studies report similar limitations, including poor OOD generalization~\citep{jin2025rl}, degradation on general tasks after math SFT~\citep{huan2025does}, and forgetting caused by off-policy supervision~\citep{lai2025reinforcement,chen2025retaining}. 

However, other studies suggest that SFT can generalize under suitable conditions. 
Long CoTs, sufficient optimization, high-quality data, and strong base models have all been identified as important factors for SFT generalization~\citep{yang2025demystifying,ren2026rethinking}. 
Moreover, \citet{tian2026learning} surprisingly find that training on incorrect CoTs can improve OOD generalization, and \citet{lin2025debunk} observes that CoT supervision with diverse prompts can even surpass RL in generalization performance. 
These findings suggest that poor SFT generalization may reflect unfavorable training conditions rather than an inherent inability.

\paragraph{Interaction between SFT and RL.}
With the success of DeepSeek-R1~\citep{guo2025deepseek}, combining SFT and RL has become a common post-training paradigm for modern reasoning models~\citep{yu2025dapo,zheng2025group,shao2024deepseekmath,jaech2024openai,yang2025qwen3}. 
A mainstream view is that SFT provides basic reasoning behaviors, while RL further refines them through exploration and feedback~\citep{yue2026does,matsutani2026rl,zhang2025interplay}. 
Under this view, \citet{zhang2025interplay} argue that post-training should place greater emphasis on SFT, with RL serving mainly as a lightweight refinement stage.

However, the benefit of SFT before RL appears to be conditional rather than universal. 
In VLM reasoning, \citet{chen2025sft} find that SFT can impair subsequent RL, and \citet{chen2026the} show that several hybrid strategies, including SFT-then-RL, interleaved training, and model merging, may underperform RL-only training. 
Similarly, \citet{yu2025reassessing} show that SFT benefits RL mainly when the base model is weak, but can become harmful when the base model is already strong.

Recent studies therefore shift the focus from whether to combine SFT and RL to how SFT should be used for RL initialization. 
Although various methods have been proposed to better integrate the two paradigms~\citep{fu2026srft,zhang2026onpolicy}, \citet{limozin2026sft} show that some reported gains over SFT-then-RL were due to bugs in the training framework; after correction, a simple SFT-then-RL pipeline achieves the best performance. 
Consistently, \citet{ding2025rethinking} argue that SFT-then-RL remains a strong post-training paradigm and find that early-stage SFT checkpoints are better starting points for subsequent RL, in line with \citet{li2026getting}.
This also suggests that stronger standalone SFT performance does not necessarily imply better RL initialization~\citep{zhang2026good}. 
To select suitable SFT checkpoints, \citet{kang2026quagmires} further propose using validation loss or Pass@k with a large-\(k\) as more reliable indicators.


\section{Challenges and Future Directions}


\paragraph{Harmonizing Reasoning Enhancement with Knowledge Learning.}
While confidence-based SFT strategies excel at boosting reasoning, they inadvertently reinforce pre-existing beliefs, suppressing learning signals for novel facts and under-represented domains~\citep{Wu2026dft}.
To overcome this vulnerability in knowledge-intensive tasks, future research should prioritize capability-aware optimization frameworks that seamlessly balance reasoning refinement with factual acquisition. 
Achieving this synergy is essential for training versatile foundation models that act as both rigorous reasoners and reliable knowledge repositories.


\paragraph{Inference-Efficient Reasoning Trajectories.}
While current reasoning-oriented SFT frameworks have achieved remarkable strides in generalization and task accuracy, this progress often inadvertently sacrifices inference efficiency.
By strictly cloning verbose expert demonstrations, models are conditioned to generate excessively long CoT trajectories, incurring severe computational overhead and elevated latency during deployment~\cite{kang2025c3ot}.
Consequently, explicitly optimizing the length-accuracy trade-off has emerged as a crucial future direction.
Future frameworks must transition from passively replicating extensive reasoning chains to actively compressing them, exploring novel mechanisms to distill prolonged logical steps into dense, condensed trajectories without compromising the underlying mathematical rigor or deductive soundness.

\paragraph{Toward RL-Compatible Supervised Fine-Tuning.}
In modern post-training pipelines, SFT is rarely used in isolation, but typically serves as a preparatory stage for RL~\citep{guo2025deepseek}.
However, as prior work has observed~\citep{zhang2026good}, better standalone SFT performance does not necessarily translate into a better RL initialization.
This suggests that existing SFT methods may have been evaluated under an incomplete criterion: most data construction and algorithmic designs focus on post-SFT performance, while overlooking their compatibility with RL.
Therefore, an important future direction is to re-evaluate existing methods under the full SFT-then-RL pipeline, examining which types of data, objectives, and checkpoints provide better entry for RL.
Beyond evaluation, future work should also develop RL-compatible SFT objectives that preserve exploration potential, policy diversity, and reward-improvable behavior.

\paragraph{Toward Causal Analysis of Reasoning SFT.}
Existing analyses have shown that factors such as CoT length, data difficulty, diversity, reasoning patterns, and data quantity can affect reasoning SFT~\citep{shen2025long,muennighoff2025s1}. However, many conclusions remain correlational and may be confounded by other variables.
For example, longer CoTs may improve generalization not only because of their length, but also because they change the reasoning structure, token budget, problem difficulty, or teacher-model distribution~\citep{yang2025demystifying,chandra2025shape}.
Future work should therefore conduct more controlled studies that isolate the causal effects of individual data properties, model capability, and training dynamics. Such analyses would lead to a clearer understanding of why and when reasoning SFT improves generalization.
\section{Conclusion}
In this survey, we present a comprehensive review of recent advancements in reasoning SFT, examining the literature through the dual lenses of methodological design and analytical investigation.
We established a structured taxonomy categorizing methods into data-centric and algorithm-centric approaches, while synthesizing fragmented findings across data, optimization dynamics, and mechanistic analysis.
By outlining key bottlenecks, we further summarize several promising future directions.
Ultimately, we hope this survey serves as a foundational roadmap to guide researchers and inspire subsequent breakthroughs in developing advanced reasoning foundation models.

\section*{Limitations}
Despite our effort to provide a comprehensive review, this survey has several limitations.
First, reasoning SFT is a rapidly evolving field, and new methods and analyses continue to emerge at a fast pace; therefore, some very recent or concurrent studies may not be fully covered.
Additionally, this survey primarily focuses on recent studies, especially those published since 2025, in order to capture the latest methodological and analytical developments in reasoning-oriented SFT.
As a result, we do not provide a systematic review of earlier works that mainly concentrated on constructing reasoning or CoT-style training data.
Although these efforts laid important foundations for reasoning SFT, a comprehensive historical account of data construction methods is beyond the scope of this survey.

\section*{Acknowledgments}
This work was supported by the New Generation Artificial Intelligence-National Science and Technology Major Project (2025ZD0123501), and the National Natural Science Foundation of China under Grants U2441251 and 62276256.
We also extend our sincere thanks to the anonymous reviewers for their constructive suggestions.

\bibliography{main}

\appendix
\newpage
\appendix
\clearpage

\section{Data-Centric Methods}
\label{sec:methods_data_1}

Reasoning SFT is shaped by how problems are sourced, reasoning trajectories are constructed and verified, and training examples are selected. We therefore organize data-centric methods according to the pipeline:

\[
\begin{aligned}
&\textit{Problem Sourcing}
\rightarrow
\textit{CoT Trajectory Synthesis} \rightarrow\\
&\textit{Trajectory Verification} \rightarrow\\
&\textit{Data Filtering and Curation}.
\end{aligned}
\]
When a work spans multiple stages, we assign it to the stage of its primary contribution.

\paragraph{Problem Sourcing.}
Human-authored datasets such as GSM8K~\cite{cobbe2021training} and MATH~\cite{hendrycks2measuring} provide reliable mathematical problems, multi-step solutions, and broad difficulty coverage, but are expensive to construct and scale. 
Synthetic resources, including principled logic corpora and ProofWriter~\cite{tafjord2021proofwriter}, generate theories, questions, and deduction chains with controlled structures~\cite{morishita2024enhancing}. 
PrOntoQA~\cite{saparov2022language} also uses synthetic logical instances, but primarily serves as a benchmark for analyzing deductive reasoning rather than as a standard SFT corpus. 
Other works mine or transform public resources: DeepSeekMath~\cite{shao2024deepseekmath} constructs a large mathematical corpus from filtered web data, NuminaMath~\cite{li2024numinamath} releases a large collection of competition-math problems and solutions, and MIND~\cite{akter2025mind} converts OpenWebMath~\cite{paster2024openwebmath} into synthetic dialogues for mathematical pretraining. 
Problem reformulation methods further expand existing distributions through question rewriting~\cite{yu2024metamath}, back-translation~\cite{lu2024mathgenie}, or distilled problem synthesis~\cite{zhou2024jiuzhang3}.

\paragraph{CoT Trajectory Synthesis.}
Trajectory synthesis converts problems into explicit reasoning demonstrations.
~\cite{cai2024system} adds reasoning plans and intermediate steps to existing answers, while STaR~\cite{zelikman2022star} bootstraps rationales from question--answer pairs with known final answers but without rationale annotations.
Teacher-model distillation and self-improvement produce large-scale reasoning data, including hybrid natural-language and program-based solutions in MAmmoTH~\cite{yue2024mammoth}, tool-integrated trajectories in ToRA~\cite{gou2024tora} and MathCoder~\cite{wang2024mathcoder}, and mathematical and coding reasoning traces in Qwen2.5-Math~\cite{yang2024qwen2} and OpenCodeReasoning~\cite{ahmad2025opencodereasoning}, respectively.
Sampling-based methods explore multiple candidate paths through search or self-evolution, as in FastMCTS~\cite{li2025fastmcts} and rStar-Math~\cite{guan2025rstar}. 
DeepSeek-Prover~\cite{xin2024deepseek} extends this approach to Lean theorem-proving trajectories.

\paragraph{Trajectory Verification.}
Verification can target final outcomes or intermediate reasoning steps. 
Exact answer matching~\cite{cobbe2021training}, code execution~\cite{gou2024tora, wang2024mathcoder}, unit tests~\cite{ahmad2025opencodereasoning}, and Lean proof checking~\cite{xin2024deepseek} provide deterministic signals for mathematical, executable, and formal solutions. 
However, execution-based filtering is not always beneficial: OpenCodeReasoning reports that discarding all execution-failing solutions can reduce downstream performance~\cite{ahmad2025opencodereasoning}.
Process-level verification is studied through human step annotations in PRM800K~\cite{lightman2024let}, automatic process supervision in Math-Shepherd~\cite{wang2024math}, and error localization in ProcessBench~\cite{zheng2025processbench}, the latter being primarily an evaluation benchmark rather than an SFT dataset. 
AceMath~\cite{liu2025acemath} and Qwen2.5-Math~\cite{yang2024qwen2} additionally use outcome-level reward models to rank complete solutions, which differs from process-level verification.

\begin{table*}[t]
\centering
\small
\caption{The 2D Map: Gradient Granularity vs. Data Source/Heuristic.}
\label{tab:2d_map}
\renewcommand{\arraystretch}{1}
\begin{tabular}{l >{\raggedright\arraybackslash}m{3.2cm} >{\raggedright\arraybackslash}m{2.5cm} >{\raggedright\arraybackslash}m{2.5cm} >{\raggedright\arraybackslash}m{2.5cm}}
\toprule
\textbf{Granularity} & \textbf{Model Confidence / Entropy} & \textbf{Sample Difficulty} & \textbf{Semantic / Structural Rules} & \textbf{Reference Policy Divergence} \\
\midrule
\textbf{Token-Level} & DFT, EAFT, DEFT, RANKTUNER, ProFit, SED-SFT & SFT-GO, TS-PEFT, TALR & SFTKey, ssToken, CFT, XTF & PEAR, AESL, GIFT, Token Cleaning, TOSS, OTR \\ \midrule
\textbf{Sample-Level} & SelectIT, GRAPE, Q-Tuning & FLOW, BFT, RSR, DC-SFT & T-SHIRT, ASLEC & iw-SFT, BOLT, ICA, LESS, CADS, ScaleBiO \\ \midrule
\textbf{Distribution-Level} & CurioSFT & \textit{(N/A)} & \textit{(N/A)} & ASFT, PSFT, Anchored Learning, GEM \\ \midrule
\textbf{RL Integration} & RED, SRFT, Prefix-RFT & LESE, ReLIFT, DYPO, AMFT, HPT & BRIDGE, CHORD, TRAPO, BREAD, SRL, DyME, UFT, and $R^3$ & LUFFY, S-GRPO \\
\bottomrule
\end{tabular}
\end{table*}

\paragraph{Data Filtering and Curation.}
The final stage determines which candidate examples enter the training corpus.
DART-Math~\cite{tong2024dart} allocates additional generation budget to difficult problems through difficulty-aware rejection tuning.
In contrast, RedStar~\cite{xu2025redstar} primarily studies the effects of Long-CoT data scale and problem difficulty rather than introducing a dedicated filtering algorithm.
s1~\cite{muennighoff2025s1} constructs a compact dataset based explicitly on quality, difficulty, and diversity, while LIMO~\cite{ye2025limo} demonstrates the effectiveness of a small collection of difficult, diverse problems paired with high-quality reasoning traces.
HoneyBee~\cite{bansal2026honeybee} investigates analogous data construction and scaling dimensions for multimodal reasoning, while OpenThoughts~\cite{guha2026openthoughts} systematically analyzes question sourcing, answer generation, filtering, deduplication, and teacher selection. 
OpenMathInstruct-2~\cite{toshniwal2025openmathinstruct} investigates large-scale mathematical instruction data, with particular attention to solution quality and question diversity, whereas NaturalThoughts~\cite{li2025naturalthoughts} studies how to select teacher-generated reasoning trajectories that transfer efficiently to student models.
Outcome-level reward models in AceMath~\cite{liu2025acemath} and Qwen2.5-Math~\cite{yang2024qwen2} can additionally score and rank complete candidate solutions.

Overall, problem sourcing determines the input distribution, trajectory synthesis determines the reasoning behaviors available for supervision, verification estimates trajectory reliability, and filtering determines the final training distribution.
Difficulty, diversity, model provenance, and reward scores are therefore criteria within the pipeline rather than independent top-level categories.

\section{A Two-Dimensional Taxonomy of Algorithm-Centric Methods}

\noindent As detailed in Section \ref{sec:methods_Objective} of the main paper, our primary taxonomy categorizes algorithm-centric methods based on their \textit{optimization granularity}---encompassing Token-Level, Sample-Level, Distribution-Level, and RL Integration. These core dimensions constitute the Y-axis of Table \ref{tab:2d_map}. To provide a more comprehensive perspective and clarify how these methods intersect with various heuristic motivations, this 2D map cross-references the optimization granularities with their underlying data sources and algorithmic heuristics (the X-axis). 

The definitions for each heuristic category along the X-axis are explicitly stated as follows:
\textbf{Model Confidence / Entropy}: Modulating optimization signals via internal probability distributions.
\textbf{Sample Difficult}y: Allocating training focus by diagnosing whether a sample aligns with the model's own distribution, rather than its general ambiguity.
\textbf{Semantic / Structural Rules}: Applying structural constraints or filters via the semantic content of trajectories.
\textbf{Reference Policy Divergence}: Guiding learning dynamics by measuring deviations against a reference policy.

\section{Integration with RL via Off-Policy Expert Demonstrations} 
\label{Integration with RL add}
Beyond joint optimization of SFT and RL objectives, a distinct research paradigm leverages offline expert demonstrations as auxiliary exploration anchors to systematically resolve the reward sparsity and imitation bottlenecks inherent in online Reinforcement Learning (RL)~\citep{xue2026reason,lu2026one,lu2025uni,yu2026understanding}.
By integrating off-policy trajectories directly into the online policy-gradient loop rather than forcing direct cross-entropy minimization, these methods bridge the gap between static imitation and autonomous exploration.

Among these approaches, LUFFY~\cite{yan2026learning} and S-GRPO~\cite{yan2026s} explicitly integrate full expert trajectories into the candidate pool of standard relative group optimization.
LUFFY~\cite{yan2026learning} introduces Mixed-Policy GRPO, which combines off-policy demonstrations with on-policy rollouts for advantage computation, employing entropic policy shaping to reinforce low-probability yet critical actions. 
Similarly, S-GRPO~\cite{yan2026s} addresses optimization collapse in visual tasks by injecting a verified ground-truth anchor into the candidate pool whenever exploratory failure is detected, ensuring positive advantage signals to break training deadlocks.

To grant the policy higher exploratory autonomy, another branch of research truncates expert demonstrations into prefixes or hints, prompting the model to explore and complete the remaining paths.
UFT~\cite{liu2026uft} and $R^3$~\cite{xi2024training} establish reverse curricula to reduce exploration difficulty. 
Specifically, $R^3$ progressively shifts the exploration starting state backwards from the demonstration's end, while UFT smoothly decays hint lengths to zero, blending log-likelihood objectives with reward optimization. 
To stabilize this hybrid learning, Prefix-RFT~\cite{huang2025blending} reinforces offline prefixes using dynamic advantages and entropy-based clipping, preventing off-policy gradients from dominating optimization. 
Advancing this, BREAD~\cite{zhang2026bread} employs Episode Anchor Search (EAS) to adaptively select branching points, providing self-paced curriculum guidance where standard SFT-then-RL pipelines stall.
Finally, to extract dense signals without rigid token-level imitation, SRL~\cite{deng2025supervised} reformulates reasoning as a sequential decision-making process of discrete logical "actions." By generating internal monologues before committing to each step, SRL computes sequence similarity rewards strictly between generated and expert actions.

\begin{table*}[t]
\centering
\small
\begin{tabularx}{\textwidth}{p{0.22\textwidth} X}
\toprule
\textbf{Task Type} & \textbf{Benchmarks} \\
\midrule
Mathematical reasoning 
& \textbf{Text-only:} GSM8K~\citep{cobbe2021training}, MATH~\citep{hendrycks2measuring}, MATH-401~\citep{yuan2023well}, AIME24/25~\citep{aim2024aime}, AMC~\citep{maa2023amc}, Olympiad Bench~\citep{he2024olympiadbench}.
\textbf{Multi-modal:} MathVista~\citep{lumathvista}, MathVision~\citep{wang2024measuring}, Geometry3K~\citep{lu2021inter}, GEOQA~\citep{chen2021geoqa}, UniGEO~\citep{chen2022unigeo}, MathVerse~\citep{zhang2024mathverse}, We-MATH~\citep{qiao2024we}. \\ \midrule

Spatial reasoning 
&\textbf{Multi-modal:} EmbSpatial-Bench~\citep{du2024embspatial}, Space3D-Bench~\citep{szymanska2024space3d}, SpatialRGPT-Bench~\citep{cheng2024spatialrgpt}, SpatialVLM~\citep{chen2024spatialvlm}, VSI-Bench~\citep{yang2025thinking}, OmniSpatial~\citep{jia2025omnispatial}, 3DSRBench~\citep{ma20253dsrbench}, MathSpatial~\citep{lu2026mllms}, Mmsi-bench~\citep{yang2025mmsi}. \\ \midrule

Scientific reasoning 
& \textbf{Text-only:} MMLU-Pro~\citep{wang2024mmlu}, GPQA~\citep{rein2024gpqa}, SuperGQPA~\citep{du2026supergpqa}, SciQ~\citep{welbl2017crowdsourcing}, 
HLE~\citep{phan2025humanity},
MatSciBench~\citep{zhang2025matscibench}, SciKnowEval~\citep{feng2024sciknoweval}.
\textbf{Multi-modal:} ScienceQA~\citep{saikh2022scienceqa}, MMMU~\citep{yue2024mmmu}, MMSciBench~\citep{ye2025mmscibench}. \\ \midrule

Logical reasoning 
& \textbf{Text-only:} LogiQA~\citep{liu2020logiqa}, AutoLogi~\citep{zhu2025autologi}, ZebraLogic~\citep{lin2025zebralogic}, ProofWriter~\citep{tafjord2021proofwriter}, FOLIO~\citep{han2024folio}, PrOntoQA~\citep{saparov2023language}, CLUTTR~\citep{sinha2019clutrr}.
\textbf{Multi-modal:} LogicVista~\citep{xiao2024logicvista}, VisuLogic~\citep{xu2025visulogic}, MuSLR~\citep{xu2026muslr}. \\ \midrule

Code reasoning 
& \textbf{Text-only:} SWE-bench~\citep{jimenez2024swe}, LiveCodeBench~\citep{jain2025livecodebench}, 
SecCodeBench~\citep{chen2026seccodebench}, APPS~\citep{hendrycks2measuringcode}, MBPP~\citep{austin2021program}, HumanEval~\citep{chen2021evaluating}, HumanEval+~\citep{liu2023your}, BigCodeBench~\citep{zhuo2025bigcodebench}, NaturalCodeBench~\citep{zhang2024naturalcodebench}. \textbf{Multi-modal:} Design2Code~\citep{si2025design2code}, ChartMimic~\citep{yang2025chartmimic}, UniSVG~\citep{li2025unisvg}. \\ 
\bottomrule
\end{tabularx}
\caption{Representative benchmarks for evaluating reasoning capabilities.}
\label{tab:reasoning_benchmarks}
\end{table*}
\section{Reasoning Benchmarks}
\label{app:benchmarks}

The rapid development of large reasoning models has been accompanied by an increasingly diverse set of benchmarks for evaluating reasoning capabilities. Unlike general instruction-following evaluations or retrieval tasks~\citep{xue2023dual,xue2025structcoh}, reasoning benchmarks emphasize whether models can decompose complex problems, perform multi-step inference, maintain logical consistency, and arrive at verifiable final answers. These benchmarks therefore serve as an essential empirical foundation for studying reasoning-oriented supervised fine-tuning: they not only quantify the effectiveness of different SFT data and objectives, but also expose limitations such as memorization, poor out-of-distribution generalization, and insufficient transfer across reasoning domains.

In this section, we summarize representative benchmarks commonly used to evaluate reasoning models in \Cref{tab:reasoning_benchmarks}. We organize them according to the type of reasoning skill they primarily assess, including mathematical reasoning, code reasoning, general and logical reasoning, scientific reasoning, and spatial reasoning. This overview is intended to provide a practical reference for interpreting experimental results in the literature and for selecting appropriate evaluation suites when developing or analyzing reasoning SFT methods.

\section{Comparison between Reasoning SFT and Non-Reasoning SFT}
In the study of LLMs, instruct tuning and Reasoning SFT represent two distinct training paradigms. Instruct Tuning focuses on enabling models to comprehend and execute a wide variety of human instructions directly; its outputs are typically final, fluent textual responses tailored to specific tasks such as creative writing, information extraction, or general Q\&A. Conversely, Reasoning SFT is dedicated to cultivating the model's deep logical thinking and problem-solving capabilities. Rather than just delivering the final answer, it trains the model to generate a structured "Chain of Thought" that explicitly details the step-by-step analysis, mathematical derivations, and self-correction processes required to tackle complex mathematical, scientific, or logical challenges.
We provide a demonstration of the two in \Cref{tab:cmp_sft}.

\begin{figure*}[htbp]
    \centering
    \small
\begin{tabularx}{\textwidth}{>{\hsize=0.6\linewidth}X | 
  >{\hsize=1.4\linewidth}X}
        \toprule
        \textbf{\textcolor{blue!80!black}{Instruction Tuning:}} & 
        \textbf{\textcolor{red!80!black}{Reasoning SFT:}} \\
        Make a persuasive argument to promote recycling. & 
        Let $f(x) = 1 + x + x^2 + \cdots + x^{100}$. Find $f'(1)$. \\
        \midrule
        Recycling is an incredibly important action that we must all take to help preserve our planet for future generations. By recycling, we can reduce the amount of waste that goes to landfills, conserve natural resources and energy, and reduce pollution from production of new materials.\newline\newline When we don't recycle, our garbage ends up in landfills where it can take hundreds of years to decompose. Landfills can release harmful chemicals and greenhouse gases, polluting the soil, air and water. However, by recycling and reusing materials, we minimize the amount of garbage that ends up in landfills, and therefore reduce the potential harm it can cause to the environment.\newline\newline Recycling also helps conserve our natural resources by reducing the demand for new materials. For example, recycling paper means fewer trees need to be cut down, and reusing metal and plastic means less mining and oil extraction. This, in turn, conserves energy, minimizes deforestation and reduces the environmental impact associated with producing new materials.\newline\newline Additionally, recycling also has a positive impact on the economy. It creates jobs in the recycling and manufacturing industries, and reduces the costs of waste disposal for companies and taxpayers. Recycling is good for the environment and the economy.\newline\newline In conclusion, recycling is an easy and effective way to take action and make a difference for the future of our planet. By simply separating our recyclable materials and disposing of them properly, we can help reduce waste, conserve resources, and protect the environment. We all have a part to play and every little action counts. So, let's make the smart choice and recycle today for a better tomorrow. & 
\textbf{Step 1: Recognize the form of the function}

This is a geometric series:

$$
f(x) = \sum_{k=0}^{100} x^k
$$

This is the sum of a geometric series with first term $ 1 $, common ratio $ x $, and $ 101 $ terms.

The sum of a finite geometric series is:

$$
f(x) = \frac{1 - x^{101}}{1 - x} \quad \text{for } x \ne 1
$$

This is a standard formula for the sum of a finite geometric series.

\textbf{Step 2: Differentiate the function}

We differentiate the expression for $ f(x) $:

$$
f(x) = \frac{1 - x^{101}}{1 - x}
$$

We will use the \textbf{quotient rule}:

$$
f'(x) = \frac{(1 - x)^2 \cdot \frac{d}{dx}(1 - x^{101}) - (1 - x^{101}) \cdot \frac{d}{dx}(1 - x)}{(1 - x)^2}
$$

Let's compute the derivatives:

- $\frac{d}{dx}(1 - x^{101}) = -101x^{100}$
- $\frac{d}{dx}(1 - x) = -1$

Now plug these into the quotient rule:

$$
f'(x) = \frac{(1 - x)^2(-101x^{100}) - (1 - x^{101})(-1)}{(1 - x)^2}
$$

Simplify the numerator:

$$
(1 - x)^2(-101x^{100}) + (1 - x^{101})
$$

Now, evaluate this at $ x = 1 $:

Note that the original expression for $ f(x) $ uses a division by $ 1 - x $, which is zero at $ x = 1 $, so the derivative formula is not directly valid at $ x = 1 $. Instead, we should \textbf{differentiate the original series directly}, which is easier.

\textbf{Step 3: Differentiate the series directly}

We have:

$$
f(x) = \sum_{k=0}^{100} x^k
$$

Differentiate term by term:

$$
f'(x) = \sum_{k=1}^{100} k x^{k-1}
$$

So:

$$
f'(1) = \sum_{k=1}^{100} k \cdot 1^{k-1} = \sum_{k=1}^{100} k = \frac{100 \cdot 101}{2} = 5050
$$

\textbf{Final Answer:}

$$
\boxed{5050}
$$ \\
        \bottomrule
    \end{tabularx}
\caption{Comparison of Instruction Tuning and Reasoning SFT.}
\label{tab:cmp_sft}
\end{figure*}

\section{Bridging Analytical Insights and Methodological Design}
\label{sec:bridging_insights}

To explicitly trace how empirical observations motivate methodological innovations, we outline four key connections between our analytical findings and the corresponding method designs:

\paragraph{Distributional Alignment Drives Data Refinement.} 
\Cref{sec:std_data} show that LLM-paraphrased CoTs outperform manual ones despite errors, indicating that distributional alignment matters more than surface correctness. These findings directly motivate the data refinement methods in \Cref{sec:methods}. Self-alignment frameworks replace expert demonstrations with the model's own in-distribution generations, explicitly optimizing for distributional alignment rather than importing off-policy trajectories.

\paragraph{Diversity Requirements Motivate Hybrid Trajectory Augmentation.} 
Studies in \Cref{sec:std_data} highlight the importance of prompt and response diversity for generalization, with purely imitative training leading to overfitting and limited robustness. This finding drives data augmentation methods such as GFT~\citep{gan2026gft}, which mitigates single-path dependency by pairing queries with hybrid ensembles of expert, teacher, and self-generated trajectories to actively enrich training diversity.

\paragraph{Overfitting Risks Motivate Selective Token Optimization.} 
\Cref{sec:std_mec} indicate that excessive SFT can compromise reasoning capabilities and lead to poor generalization. These risks motivate token-level interventions that constrain the scope and intensity of SFT updates. Token-masking strategies like SFTKey~\citep{shi2025rethinking} restrict optimization to answer tokens. Similarly, reweighting methods such as DEFT~\citep{wang2026gradients} and VCORE~\citep{gong2025vcore} modulate per-token gradients to prioritize high-entropy or high-loss tokens, selectively limiting the degree of behavioral change imposed by SFT.

\paragraph{SFT-RL Complementarity Drives Integration Methods.} 
\Cref{sec:std_mec} reveal that SFT expands the model's capability boundary (Pass@k), while RL refines solution quality through exploration. This complementary relationship motivates the SFT+RL integration methods: SRFT~\citep{fu2026srft} preserves exploration potential via entropy-regularized SFT, and CHORD~\citep{zhang2026onpolicy} schedules smooth transitions from SFT to RL.

\vspace{1em}
\noindent These four connections demonstrate how empirical findings expose fundamental bottlenecks, which in turn motivate specific methodological innovations. By making these causal links explicit, we explain not only what methods exist, but why they were developed and how they address observed phenomena.

\section{A Unified Taxonomy of Optimization Objectives}
To deeply dissect the structural disparities across various optimization paradigms, Table~\ref{table: Objective token reweighting}, Table~\ref{table: Objective token selection}, Table~\ref{table: Objective sample reweighting}, Table~\ref{table: Objective sample selection}, Table~\ref{table: Objective distribution-level regularization}, and Table~\ref{table: Objective fusion with RL} systematically formalize the loss functions and gradient expressions of contemporary reasoning-oriented SFT methodologies. 
By unifying these disparate formulations into a standardized mathematical framework, we crystallize the evolutionary trajectory that spans from fine-grained gradient adjustments (\textit{Token/Sample-Level}) to macroscopic policy constraints (\textit{Distribution-Level}). 
This taxonomy not only provides an intuitive quantitative baseline for comparing diverse optimization objectives, but also enables readers to clearly demarcate different paradigms based on their data utilization strategies (\textit{offline vs. online}) and the mechanics of their SFT-RL integration.
\begin{sidewaystable*}[h]
\centering
\caption{Objective comparison across training paradigms for token reweighting methods.}
\label{table: Objective token reweighting}
\scriptsize
\renewcommand{\arraystretch}{3.0}
\begin{tabularx}{\textwidth}{@{}m{0.1\textwidth}m{0.1\textwidth}m{0.73\textwidth}@{}}
\toprule
\midrule
\textbf{Method} & \textbf{Paper} & \textbf{Objective Expression}  \\
\midrule
DFT & \cite{Wu2026dft} &  $\mathcal{L}_{\text{DFT}} = \mathbb{E}_{(x, y^\star) \sim \mathcal{D}} \left[ -\sum_{t=1}^{|y^\star|} \text{sg} \left( \pi_\theta(y_t^\star \mid y_{<t}^\star, x) \right) \log \pi_\theta(y_t^\star \mid y_{<t}^\star, x) \right]$\\
VCORE &  \cite{gong2025vcore} & 
\makecell[l]{  $\mathcal{L}_{\text{VCORE}} = \sum_{(x,y) \in \mathcal{B}} \left[ \sum_{t \ge 1} \frac{\exp(\tau s_t(x, y, \theta))}{\sum_{j \ge 1} \exp(\tau s_j(x, y, \theta))} \nabla_\theta \ell_t(\theta; x, y) \right],$  where $s_t(x, y, \theta) \triangleq \langle \nabla_\theta \mathcal{L}(\theta), \nabla_\theta \ell_t(\theta; x, y) \rangle$}  \\
EAFT & \cite{diao2026entropy} & 
\makecell[l]{$\mathcal{L}_{\text{EAFT}} = -\sum_{t=1}^{T} \frac{H_t^{\text{top-20}}}{3.0} \cdot \log P_\theta(y_t \mid \boldsymbol{x}, \boldsymbol{y}_{<t}),$ where $H_t = -\sum_{v \in \mathcal{V}} P_\theta(v \mid \boldsymbol{x}, \boldsymbol{y}_{<t}) \log P_\theta(v \mid \boldsymbol{x}, \boldsymbol{y}_{<t})$, $H_t^{\text{top-}K}$ is the entropy of the top-$K$ probability distribution.}.  \\
DEFT & \cite{wang2026gradients} & 
\makecell[l]{$\mathcal{L}_{\text{DEFT}}(\theta) = \mathbb{E}_{(c, \tilde{y}) \sim \mathcal{T}} \left[ \frac{1 - p_{\theta}(\tilde{y} | c)^{\alpha_{\text{DEFT}}(c)}}{\alpha_{\text{DEFT}}(c)} \right]$,  where $\alpha_{\text{DEFT}}(c) \triangleq \sum_{\nu \in \mathcal{V}} P_{\theta}(\nu | c)^2 \in (0, 1]$} \\
RANKTUNER & \cite{yu2026probability} & 
\makecell[l]{$\mathcal{L}(\theta) = \mathbb{E}_{(x, y) \sim \mathcal{D}} \left[ -\sum_{t=1}^{T} ( p_t \cdot \mathcal{S}_t) \log p_t \right]$,  where $\mathcal{S}_t \approx (p_t \cdot s(H_t))^{-K(\xi_t)}$, $K(\xi_t) = [\log_2(\xi_t + 1)]^{-2}$, $\xi_t = \max(R_t, s(H_t))$,  $R_t$ {is the ground-truth token's rank} } \\ 
TALR & \cite{lin2025sft} & 
\makecell[l]{$\mathcal{L}_{\text{TALR}}(\theta) = -\mathbb{E}_{(x, y) \sim \mathcal{D}} \left[ \sum_{t=1}^{|y|} w(c_t, y_t) \log P_\theta(y_t \mid c_t) \right]$,  where $w(c_t, y_t) = \left\| \nabla_\theta \log P_\theta(y_t \mid c_t) \right\| \cdot \exp\left( -\lambda \cdot \text{KL}(P_\theta(\cdot \mid c_t) \| Q_{\text{base}}(\cdot \mid c_t)) \right)$} \\
AESL & \cite{li2026getting} & 
$\mathcal{L}_{\text{AESL}}(\theta) = -\mathbb{E}_{q, s^* \sim \mathcal{D}_{\text{SFT}}} \left[ \sum_{t=1}^{|s^*|} p(q, s_t^*, \pi_\theta) \cdot \log \pi_\theta(s_t^* \mid q, s_{<t}) \right]$, where $p(q, s_t^*, \pi_\theta) = 1 - \text{softmax} \left[ \frac{y(s_t^* \mid q, s_{<t})}{-t_{\text{scaling}} \cdot \bar{l}_t} \right]$, $\bar{l}_t = \frac{1}{t} \sum_{i=1}^t -\log \pi_\theta(s_i^* \mid q, s_{<i})$ \\ 
GIFT  & \cite{zhao2026gift} &  
$\mathcal{L}_{\text{GIFT}}(\phi) = \mathbb{E}_{(x,y) \sim \mathcal{D}} \left[ \sum_{t=1}^{T} q_t \cdot \ell_t(\phi) \right] \quad $, where $\ell_t(\phi) = -\log p_{\theta_B, \phi}(y_t \mid x, y_{<t}) \quad $, $q_t = p_{\theta_I}(y_t \mid x, y_{<t}), \quad t=1,\dots,T \quad $ \\
PEAR  & \cite{zhang2026good} &   $\mathcal{L}_{\text{PEAR}}(\theta) = \mathbb{E}_{(x,y) \sim \mathcal{D}} \left[ \sum_{t=1}^{T} \text{sg}[\hat{G}_t] \cdot l_\theta(x, y_{<t}, y_t) \right] $,  where $G_t = w_{1:T} = \prod_{j=1}^{T} \Delta_j \quad \text{($\forall t=1,\dots,T$)}$ \\
\bottomrule
\end{tabularx}
\end{sidewaystable*}

\begin{sidewaystable*}[h]
\centering
\caption{Objective comparison across training paradigms for token selection methods.}
\label{table: Objective token selection}
\scriptsize
\renewcommand{\arraystretch}{3.0}
\begin{tabularx}{\textwidth}{@{}m{0.1\textwidth}m{0.1\textwidth}m{0.73\textwidth}@{}}
\toprule
\midrule
\textbf{Method} & \textbf{Paper} & \textbf{Objective Expression}  \\
\midrule
SFT-GO & \cite{kim2025sft} & $\mathcal{L}_{\text{SFT-GO}}(\theta) = (1 - \lambda)\ell_{\text{SFT}}(\theta) + \lambda L_{\text{worst}}(\theta)$, where  $\ell_{\text{SFT}}(\theta) = -\frac{1}{N} \sum_{t=1}^{N} \log \pi_\theta(y_t \mid x, y_{<t})$. Grouping tokens into $g \in \mathcal{G}$ by importance.  \\ 
TS-PEFT & \cite{ma2025ts} &
$\text{Output} = Y_{\text{backbone}} + Y_{\text{adapter}}$, $Y_{\text{adapter}} = (M \odot X) \cdot A \cdot B$, $I_t = \text{Selector}(X_t)$, the selector evaluates token importance scores based on metrics such as hidden state norms or cumulative attention scores. \\ 
Token Cleaning & \cite{pang2025token} & $\mathcal{L}_{\text{Cleaning}}(\theta) = -\frac{1}{\sum M_t} \sum_{t=1}^{N} M_t \cdot \log \pi_\theta(y_t \mid x, y_{<t})$, 
 $M_t = \begin{cases} 1, & \text{if } I_t \geq \tau \quad  \\ 0, & \text{if } I_t < \tau \quad  \end{cases}$,  where $I_t = \log\mathbb{P}(x|x;\theta^{\prime}) - \log\mathbb{P}(x|x;\theta) $, where $\theta$ represents the base model while $\theta^{\prime}$ serves as a task-aware reference model. \\ 
TOSS & \cite{li2026token} & $\mathcal{L}_{\text{TOSS}}(\theta) = -\frac{1}{\sum M_t} \sum_{t=1}^{N} M_t \cdot \log \pi_\theta(y_t \mid x, y_{<t})$, $M_t = \begin{cases} 1, & \text{if } S_t < \tau \\ 0, & \text{if } S_t \geq \tau   \end{cases}$, where $S_t = \text{Evaluator}(y_t \mid x, y_{<t})$ serves as a safety probe based on a pre-trained model.  \\
SSTOKEN & \cite{qin2025sstoken} & $\mathcal{L}_{\text{ssToken}}(\theta) = -\frac{1}{\sum_{t=1}^{T} w_t} \sum_{t=1}^{T} w_t \cdot \log P_\theta(y_t \mid x, y_{<t})$, $w_t = \sigma \left( \frac{S_t - \tau}{\tau} \right) = \frac{1}{1 + \exp\left( -\frac{S_t - \tau}{\tau} \right)}$, where $S_t = \text{MLP}\left( \text{LayerNorm}\left( \mathbf{h}_t^{(L)} \right) \right)$, where $\mathbf{h}_t^{(L)}$ is the raw hidden state vector at the final layer at token position $t$. \\
CFT & \cite{ruan2025enhancing} & $\mathcal{L}_{\text{CFT}}(\theta) = -\frac{1}{\sum_{t=1}^{T} w_t} \sum_{t=1}^{T} w_t \cdot \log P_\theta(y_t \mid x, y_{<t})$, $M_t = \begin{cases} 1, & \text{if } S_t < \tau \\ 0, & \text{if } S_t \geq \tau   \end{cases}$, where $S_t = \text{D}_{\text{KL}}\left( P_\theta(\cdot \mid x, y_{<t}) \parallel P_{\theta, \text{perturbed}}(\cdot \mid x, y_{<t}) \right)$ \\
XTF & \cite{yang2026explainable} & 
\makecell[l]{$\mathcal{L}_{XTF} = -\sum_{O_{k} \notin D_{\text{noise}}} \log P(O_{k} \mid I + [O_{0}, O_{1}, \dots, O_{k-1}])$, 
$D_{\text{noise}} = (D_{\text{RI}}) \cup (D_{\text{KN}_{\downarrow}}) \cup (D_{\text{TR}_{\downarrow}})$, 
$\mathcal{D}_{\text{RI}}(O_{k}) = \mathcal{A}(\theta, I + O)[l_{I} + k]$, \\
$\mathcal{D}_{\text{KN}}(O_{k}) = 1 - P(O_{k} \mid I + [O_{0}, O_{1}, \dots, O_{k-1}])$,
$\mathcal{S}_{\text{TR}}(O_{k}) = 1 - \text{Normalize}\Big(D\big(\mathcal{E}(O_{k}), \frac{\sum(\mathcal{E}(\theta, exp_{w}))}{n_{w}})\big)\Big)$.} \\
ProFit & \cite{liu2026profit} &  
$\mathcal{L}_{\text{ProFit}}(\theta) = \mathbb{E}_{(x, y^{*}) \sim \mathcal{D}} \left[ -\sum_{t=1}^{|y^{*}|} \mathcal{M}_{t} \cdot \log \pi_{\theta}(y_{t}^{*} \mid x, y_{<t}^{*}) \right]$, $\mathcal{M}_{t} = \mathbb{I}\Big[sg\big(\pi_{\theta}(y_{t}^{*} \mid x, y_{<t}^{*})\big) > \tau\Big]$ \\
SED-SFT & \cite{chen2026sed} & 
$\mathcal{L}_{\text{SED-SFT}}(\theta) = \sum_{t=1}^{|y^{*}|} \left[ -\log \pi_{\theta}(y_{t}^{*} \mid x, y_{<t}^{*}) + \lambda \cdot M_{t} \cdot \mathcal{L}_{\text{DE}}\big(\pi_{\theta}(y_{t}^{*} \mid x, y_{<t}^{*})\big) \right]$, $\mathcal{L}_{\text{DE}}(p) = \left( p - 0.5 \right)^{2}$, $M_{t} = \mathbb{I}[P_{\text{Top-k}}(t) < \tau]$\\
OTR & \cite{ming2025one} &  $\mathcal{L}_{\text{OTR}}(\theta) = \frac{1}{T} \sum_{t=1}^{T} \mathcal{L}_{\text{OTR}}^{t}(\theta)$, $\mathcal{L}_{\text{OTR}}^{t}(\theta) = -\frac{N_{\text{gt}}}{K} \log \pi_{\theta}(x_{t} \mid s_{t}) - \frac{\beta}{K} \sum_{j \text{ s.t. } a_{t,j}^{\prime} \neq x_{t}} \log \pi_{\theta}(a_{t,j}^{\prime} \mid s_{t})$, $R(a_{t,j}^{\prime}, x_{t}) = \begin{cases} 1 & \text{if } a_{t,j}^{\prime} = x_{t}, \\ \beta & \text{if } a_{t,j}^{\prime} \neq x_{t}. \end{cases}$ \\
SFTKey & \cite{shi2025rethinking} & 
\makecell[l]{First
 stage: $\mathcal{L}_{\text{SFT}}(\theta) = -\sum_{i=1}^{N}\sum_{t=1}^{L_{i}}\log P(y_{i,t} \mid x_{i}, y_{i,<t}; \theta)$, $y_{i,t}$ includes both the CoT steps and the final answer.\\
Second stage:$\mathcal{L}_{\text{Answer}}(\theta) = -\sum_{i=1}^{N}\sum_{t=T_{i}+1}^{L_{i}}\log P(y_{i,t} \mid x_{i}, y_{i,<t}; \theta)$, $y_{i,t}$ excludes the CoT steps and contains only the final answer.} \\
\bottomrule
\end{tabularx}
\end{sidewaystable*}

\begin{sidewaystable*}[h]
\centering
\caption{Objective comparison across training paradigms for sample reweighting methods.}
\label{table: Objective sample reweighting}
\scriptsize
\renewcommand{\arraystretch}{3.0}
\begin{tabularx}{\textwidth}{@{}m{0.1\textwidth}m{0.1\textwidth}m{0.73\textwidth}@{}}
\toprule
\midrule
\textbf{Method} & \textbf{Paper} & \textbf{Objective Expression}  \\
\midrule
FLOW & \cite{sanyal2025upweighting} & $\mathcal{L}_{FLOW}(\theta) = \sum_{i=1}^{n} w_{i} \cdot f_{i}(\theta)$ , $w_{i} = \exp\left(-\frac{f_{i}(\theta^{*})}{\tau}\right)$, where $f_i(\theta^*)$ denotes the cross-entropy loss of the $i$-th sample computed on the pre-trained model. \\
BFT & \cite{tang2025aligning} & 
\makecell[l]{$\mathcal{L}_{\text{BFT}}(\theta) = \frac{1}{B} \sum_{b=1}^{B} s_b \cdot \frac{\sum_{t} m_{b,t} w_{b,t} l_{b,t}}{\sum_{t} m_{b,t} + \epsilon}$, where $l_{b,t} = -\log \pi_{\theta}(y_{b,t} \mid y_{b,<t}, x_{b})$ , $p_b^{\text{conf}} = \min_{G_i \in \mathcal{G}_b} C_{G_i}, \quad \text{and} \quad s_b = sg(1 - p_b^{\text{conf}})$, \\ $w_{b,t} = sg\big(\pi_{\theta}(y_{b,t} \mid y_{b,<t}, x_{b})\big) = sg(c_{b,t})$} \\
iw-SFT & \cite{qin2025supervised} &  $\mathcal{L}_{\text{iw-SFT}}(\theta) = -\mathbb{E}_{\tau \in \mathcal{D}^{+}} \left[ W(\tau) \cdot \log p(\tau; \theta) \right]$, where $W(\tau) = \frac{q(\tau)}{\pi_{\text{ref}}(\tau)} = \exp \left( \sum_{i} \Big( \log \pi_{q}(a_{i} \mid s_{i}) - \log \pi_{\text{ref}}(a_{i} \mid s_{i}) \Big) \right)$ \\
BOLT & \cite{shu2026reference} & $\mathcal{L}_{\text{BOLT}}(\theta) = -\frac{1}{|\mathcal{B}|}\sum_{(x,y,w) \in \mathcal{B}} w \cdot \log \pi_{\theta}(y \mid x)$, where ${w}(x,y_{n}) = \frac{\exp(r(x,y_{n})/\beta)}{\hat{Z}_{N}(x)}, \quad \hat{Z}_{N}(x) = \frac{1}{N}\sum_{n=1}^{N}\exp(r(x,y_{n})/\beta)$ \\
ICA & \cite{zhang2025holdout} & $\mathcal{L}_{\text{ICA}}(\theta) = \mathbb{E}_{z \sim \mathcal{D}} \left[ w(z) \cdot \mathcal{L}_{\text{base}}(z; \theta) \right]$, where $w(z) = \sigma \left( \frac{\text{Score}_{\text{ICA}}(z) - \mu}{\tau} \right)$, $\text{Score}_{\text{ICA}}(z) = \sum_{z_h \in \mathcal{H}} \left( \log P_{\theta}(z_h) - \log P_{\theta}(z_h \mid z) \right)$ \\
ScaleBiO & \cite{pan2025scalebio} &  $\mathcal{L}_{\text{ScaleBiO}} =  L_{1}(\lambda, w) + L_{1}(\lambda, u) + \alpha \cdot \Big( L_{2}(\lambda, w) - L_{2}(\lambda, u) \Big)$, where $L_{1}(\lambda, w) = -\frac{1}{|\mathcal{D}_{\text{train}}|} \sum_{i \in \mathcal{D}_{\text{train}}} \lambda_{i} \cdot \log P_{w}(y_{i} \mid x_{i})$ and $ L_{2}\big(\lambda, w^{*}(\lambda)\big) = -\frac{1}{|\mathcal{D}_{\text{val}}|} \sum_{j \in \mathcal{D}_{\text{val}}} \log P_{w^{*}(\lambda)}(y_{j} \mid x_{j})$ \\
\bottomrule
\end{tabularx}
\end{sidewaystable*}

\begin{sidewaystable*}[h]
\centering
\caption{Objective comparison across training paradigms for sample selection methods.}
\label{table: Objective sample selection}
\scriptsize
\renewcommand{\arraystretch}{3.0}
\begin{tabularx}{\textwidth}{@{}m{0.1\textwidth}m{0.1\textwidth}m{0.73\textwidth}@{}}
\toprule
\midrule
\textbf{Method} & \textbf{Paper} & \textbf{Objective Expression}  \\
\midrule
SelectIT & \cite{liu2024selectit} & $ Selected sample: M(X, Y) = \mathbb{I}\Big[ \alpha_{\text{min}} \le \text{Unc}(Y \mid X) \le \alpha_{\text{max}} \Big] \wedge \mathbb{I}\Big[ S_{\text{ref}}(X, Y) \ge \tau \Big]$ , where $\text{Unc}(Y \mid X) = -\frac{1}{T} \sum_{t=1}^{T} \log P_{\theta}(y_t \mid X, y_{<t})$. $S_{\text{ref}}$ is the quality score generated by the model through self-reflection based on the prompt, and $\tau$ denotes the quality threshold. \\
GRAPE & \cite{zhang2026best} & $\mathcal{L}_{\text{GRAPE}}(\theta) = -\mathbb{E}_{(x, y^*) \sim \mathcal{D}_{\text{GRAPE}}} \left[ \sum_{t=1}^{|y^*|} \log P_{\theta}(y^*_t \mid x, y^*Ref_{<t}) \right]$, where $y^* = \arg\max_{y \in \mathcal{Y}_x} S(y \mid x), \quad \mathcal{D}_{\text{GRAPE}} = \{(x, y^*)\}$, $S(y \mid x) = \exp \left( \frac{1}{|Y|} \sum_{t=1}^{|Y|} \log P_{\theta_0}(y_t \mid x, y_{<t}) \right)$ \\
RSR & \cite{yang2026reasoning} & Selected sample: $ M(X, Y) = \mathbb{I}\Big[ \text{RSR}(Y \mid X) \ge \tau \Big], \text{RSR}(Y \mid X) = \frac{1}{T} \sum_{t=1}^{T} \text{RSR}_t, $ where $\text{RSR}_t = \frac{-\log P_{\theta}(y_t \mid X, y_{<t})}{\log \text{Rank}_{\theta}(y_t \mid X, y_{<t}) + \epsilon}$ \\
DC-SFT & \cite{lu2026does} & $ Selected sample: \mathcal{D}_{curated} = \mathcal{D}_{Easy} \cup \mathcal{D}_{Medium} = \{ (x, y) \in \mathcal{D} \mid \text{Correct}(x) > 0 \}$ \\
LESS & \cite{xia2024less} & $\text{Score}(z) = \max_{j} \, \text{Inf}_{Adam}(z, \mathcal{D}_{val}^{(j)})$, $\text{Inf}_{Adam}(z, \mathcal{D}_{val}^{(j)}) = \sum_{i=1}^{N} \overline{\eta}_i \cdot \text{cos}\left( \overline{\nabla}l(\mathcal{D}_{val}^{(j)}; \theta_i), \, \tilde{\Gamma}(z, \theta_i) \right)$, $\Gamma(z, \theta_i) = \frac{m_i}{\sqrt{v_i} + \epsilon}$ \\
CADS & \cite{wan2026computational} &  $\mathcal{L}_{CADS}^{\alpha}(\theta, s) = \mathbb{E}_{m \sim p(m|s)} \left[ \mathcal{L}_{val}(\theta) + \alpha \Big( \mathcal{L}_{trn}(\theta, m) - l(|m|) \Big)^2 \right]$, $\nabla_{s}\mathcal{L}_{CADS}^{\alpha}(\theta, s) = \mathbb{E}_{m \sim p(m|s)} \left[ \Big( \mathcal{L}_{val}(\theta) + \alpha (\mathcal{L}_{trn}(\theta, m) - l(|m|))^2 \Big) \cdot \nabla_{s}\log p(m|s) \right] $ \\
T-SHIRT & \cite{fu2026t} &
\makecell[l]{$S\text{-}IFD_{k}(x, y) = \exp \left\{ -\frac{1}{\sum_{t=1}^{T} w_t} \sum_{t=1}^{T} w_t \cdot \Delta_t \right\}$, $\text{where } w_t = \begin{cases} 1 & \text{if } \Delta_t \text{ ranks in the top } k\% \text{ of the entire dataset,} \\ 0 & \text{otherwise} \end{cases}$, $\Delta_t = \log P_{\theta'}(y_t \mid y_{<t}, x) - \log P_{\theta'}(y_t \mid y_{<t})$. \\
$\mathcal{S}_{can} = \arg\max_{\mathcal{D}, \gamma b} \, \hat{\mu}(x,y) $,  $\mathcal{S} = \arg\min_{\mathcal{S}_{can}, b} \, \hat{\sigma}^{2}(x,y) $, where $\hat{\mu}(x,y) = \frac{1}{M}\sum_{i=1}^{M} S\text{-}IFD_{k}(x^{(i)\prime}, y^{(i)\prime})$, $\hat{\sigma}^{2}(x,y) = \frac{1}{M}\sum_{i=1}^{M}\left(S\text{-}IFD_{k}(x^{(i)\prime}, y^{(i)\prime}) - \hat{\mu}(x,y)\right)^{2}$. } \\
Q-Tuning & \cite{wang2025winning} &
$Ent(x, y; \theta) = \frac{1}{|T|} \sum_{i \in T} H(\pi_{\theta}(y_i | y_{<i}, x))$, $PPL(x, y; \theta) = \exp\left( \frac{1}{|T|} \sum_{i \in T} -\log \pi_{\theta}(y_i | y_{<i}, x) \right)$,   only Keep samples with high entropy. \\

\bottomrule
\end{tabularx}
\end{sidewaystable*}

\begin{sidewaystable*}[h]
\centering
\caption{Objective comparison across training paradigms for distribution-level regularization.}
\label{table: Objective distribution-level regularization}
\scriptsize
\renewcommand{\arraystretch}{3.0}
\begin{tabularx}{\textwidth}{@{}m{0.1\textwidth}m{0.1\textwidth}m{0.73\textwidth}@{}}
\toprule
\midrule
\textbf{Method} & \textbf{Paper} & \textbf{Objective Expression}  \\
\midrule
ASFT & \cite{zhu2025anchored} & $\mathcal{L}_{ASFT}(\theta) = \mathcal{L}_{DFT}(\theta) + \lambda \cdot D_{KL}\left( \pi_{\theta} \,||\, \pi_{anchor} \right)$ \\
PSFT) & \cite{zhu2025proximal} &  $\mathcal{L}_{PSFT}(\theta) = -\mathbb{E}_{(x, y) \sim \mathcal{D}} \left[ \min \left( r_t(\theta) \cdot A(x, y), \, \text{clip}(r_t(\theta), 1-\epsilon, 1+\epsilon) \cdot A(x, y) \right) \right]$, where $r_t(\theta) = \frac{\pi_{\theta}(y|x)}{\pi_{\theta_{old}}(y|x)}$ \\
Anchored Learning & \cite{wang2026stabilizing} & 
$\theta^{(t+1)} = \arg \min_{\theta} \mathcal{L}_{distill}(\theta; q^{(t)}) \approx \arg \min_{\theta} \mathbb{E}_{(x,y)\sim\mathcal{D}} [KL(q^{(t)}(\cdot|x) || p_{\theta}(\cdot|x))]$, where $q^{(t)} = (1-\alpha)p_{\theta^{(t)}} + \alpha p_{sft}$ \\
GEM & \cite{li2025preserving} & $\mathcal{L}_{GEM}(\theta) = \sum_{i} \sum_{t=1}^T \sum_{y^{gene}} q_k(y^{gene}|x_i, y_{1:t-1}^{real}) \cdot \left[ \log f_\theta(y^{gene}|x_i, y_{1:t-1}^{real}) - \log f_\theta(y_i^{real}|x_i, y_{1:t-1}^{real}) \right]$, where $q = \text{softmax}\left(\frac{1}{\beta} \cdot \log f\right)$ \\
CurioSFT & \cite{wang2026learning} &
$\mathcal{L}_{total}(\theta) = \mathcal{L}_{SFT}(\theta) + \alpha \cdot \mathcal{L}_{SED}(\theta)$, where $\mathcal{L}_{SED}(\theta) = \frac{1}{2}\sum_{t=1}^{T}\left(\log\frac{\pi_{\theta}(y_t|s_t; \tau=1.0)}{\pi_{\phi}^{tch}(y_t|s_t; \hat{\tau}_t)}\right)^{2}$ \\
\bottomrule
\end{tabularx}
\end{sidewaystable*}

\begin{sidewaystable*}[h]
\centering
\caption{Objective comparison across training paradigms for integration with RL.}
\label{table: Objective fusion with RL}
\scriptsize
\renewcommand{\arraystretch}{3.0}
\begin{tabularx}{\textwidth}{@{}m{0.1\textwidth}m{0.1\textwidth}m{0.73\textwidth}@{}}
\toprule
\midrule
\textbf{Method} & \textbf{Paper} & \textbf{Objective Expression}  \\
\midrule
LESE & \cite{yang2026learning} & $L = (1 - \alpha) \cdot L_{\text{SFT}} + \alpha \cdot L_{\text{GRPO}}$, $ \alpha = \begin{cases} 
\min \left(1, l \cdot (e^M - 1) \cdot \frac{D}{D_{\text{max}}} \right), & \text{if } M \ge c \\ 0, & \text{if } M < c \end{cases} $, $M = \frac{1}{G} \sum_{i=1}^{G} \mathbb{I}(R_{\text{core}}(o_i) \ge R_{\text{thres}})$, $D = \text{StdDev} \left( \left\{ \sum_{j=1}^{K} R_{j,i} \right\}_{i=1}^{G} \right)$, $D_{\text{max}}$ is the theoretical upper bound of the standard deviation of in-group rewards.\\
HPT & \cite{lv2025towards} &
$\mathcal{L} = \alpha \mathcal{L}_{\text{RL}} + \beta \mathcal{L}_{\text{SFT}}$, $(\alpha, \beta) = \begin{cases} (0, 1), & \text{if } P \le \gamma \\ (1, 0), & \text{if } P > \gamma \end{cases}$, where $P$ is the mean pass rate of $n$ parallel rollouts for the current question. \\
AMFT & \cite{he2025amft} &
$L = (1 - \mu) \cdot L_{\text{RL}}(\theta) + \mu \cdot L_{\text{SFT}}(\theta)$, where
$\mu_{t+1} = \text{clip}\left(\mu_{t} + \eta_{\mu} \nabla_{\mu}U(\theta_{t}) + \eta_{H}(H^* - H(\pi_{\theta_{t}})), \mu_{\text{min}}, \mu_{\text{max}}\right)$, $\nabla_{\mu}U(\theta_{t}) \approx -\alpha_{\theta} \nabla_{\theta}U(\theta_{t})^{\top} \left( \nabla_{\theta}L_{\text{SFT}}(\theta_{t-1}) - \nabla_{\theta}L_{\text{RL}}(\theta_{t-1}) \right)$, $H(\pi_{\theta_t}) = \mathbb{E}_{x \sim \mathcal{D}_{SFT}} [H(\pi_{\theta_t}(\cdot|x))]$, $H^* = \mathbb{E}_{x \sim \mathcal{D}_{SFT}} [H(\pi_{\theta_W}(\cdot|x))]$. \\
CHORD & \cite{zhang2026onpolicy} & 
$\mathcal{L} = (1 - \mu)\mathcal{L}_{\text{GRPO}}(\theta) + \mu\mathcal{L}_{\text{SFT-}\phi}(\theta)$, $\mathcal{L}_{\text{SFT-}\phi}(\theta) = -\mathbb{E}_{(x, y^*) \sim \mathcal{D}_{\text{SFT}}} \left[ \sum_{t=1}^{|y^*|} \phi(y_t^*; \pi_\theta) \cdot \log \pi_\theta(y_t^* | x, y_{<t}^*) \right]$, $\phi(y_t^*; \pi_\theta) = p_t(1 - p_t)$. \\
BRIDGE & \cite{chen2025beyond} & $\overline{\theta} = (\theta, w)$, where $\theta$ is the base model parameters and $w$ denotes LoRA parameters. $\mathcal{L}_{\text{Joint}}(\theta; w) = -(1 - \lambda)J_{\text{SFT}}(\theta, w) - \lambda J_{\text{RL}}(\theta, w)$, $\mathcal{L}_{\text{Gain}}(w) = -(1 - \lambda)J_{\text{SFT}}(\theta, w) - \lambda \left[ J_{\text{RL}}(\theta, w) - J_{\text{RL}}(\hat{\theta}, w) \right]$ . $J_{\text{SFT}}$ denotes the supervised fine-tuning objective and $J_{\text{RL}}$ represents the outcome-based reinforcement learning objective. \\
SRFT & \cite{fu2026srft} &  $\mathcal{L}_{\text{SRFT}}(\theta) = \mathcal{L}_{\text{SFT}}^{\text{demo}}(\theta) + \mathcal{L}_{\text{RL}}^{\text{demo}}(\theta) + \mathcal{L}_{\text{RL}}^{\text{self-rollout}}(\theta)$, where $\mathcal{L}_{\text{SFT}}^{\text{demo}}(\theta) = w_{\text{SFT}} \cdot \mathbb{E}_{(x, y) \sim \mathcal{D}_{\text{demo}}} [-\log \pi_{\theta}(y|x)]$, $\mathcal{L}_{\text{RL}}^{\text{demo}}(\theta) = -\mathbb{E}_{(x, y) \sim \mathcal{D}_{\text{demo}}} \left[ \sum_{t=1}^{|y|} r_{k,t}(\theta) \cdot \hat{A}_k \right]$, $w_{\text{SFT}} = 0.5 \cdot \text{stop\_grad}\left(e^{-\mathcal{H}(\pi_{\theta})}\right)$, $\mathcal{L}_{\text{RL}}^{\text{self-rollout}}(\theta) = w_{\text{RL}} \cdot \mathbb{E}_{x \sim \mathcal{D}, y^+ \sim \pi_{\theta}} [-\log \pi_{\theta}(y^+|x)] + \mathbb{E}_{x \sim \mathcal{D}, y^- \sim \pi_{\theta}} [\log \pi_{\theta}(y^-|x)]$, $w_{\text{RL}} = 0.1 \cdot \text{stop\_grad}\left(e^{\mathcal{H}(\pi_{\theta})}\right)$. \\
ReLIFT & \cite{ma2025learning} &
$\mathcal{L}_{\text{GRPO}}(\theta) = -\frac{1}{G} \sum_{i=1}^{G} \frac{1}{|o_i|} \sum_{t=1}^{|o_i|} \min \left[ r_{i,t}(\theta)\hat{A}_i, \text{clip}(r_{i,t}(\theta), 1-\epsilon, 1+\epsilon)\hat{A}_i \right]$, $\mathcal{D}_{\text{hard}} = \left\{ q \ \middle| \ \max_{i \in [1, G]} R(o_i|q) = 0 \right\}$, $\mathcal{L}_{\text{SFT}}^{\text{hard}}(\theta) = -\frac{1}{|\tau^*|} \sum_{t=1}^{|1\tau^*|} \log \pi_\theta(\tau^*_t | q, \tau^*_{\le t}), \quad \text{where } q \in \mathcal{D}_{\text{hard}}$ \\
DYPO & \cite{zhu2026bridging} &
$\mathcal{L} = \sum_{q \in \mathcal{B}} \left[ \mathbb{I}(\text{All-Wrong}) \cdot \mathcal{L}_{\text{SFT}}^{\text{Multi-Teacher}}(\theta; q) + \mathbb{I}(\text{Any-Correct}) \cdot \mathcal{L}_{\text{RL}}^{\text{IS}}(\theta; q) \right]$, $\mathcal{L}_{\text{SFT}}^{\text{Multi-Teacher}}(\theta; q) = -\sum_{m=1}^{M} w_m \cdot \left[ \frac{1}{|\tau_m^*|} \sum_{t=1}^{|\tau_m^*|} \log \pi_\theta(\tau_{m,t}^* \mid q, \tau_{m,<t}^*) \right]$, $\mathcal{L}_{\text{RL}}^{\text{IS}}(\theta; q) = -\frac{1}{G} \sum_{i=1}^{G} \frac{1}{|o_i|} \sum_{t=1}^{|o_i|} \min \left[ \bar{r}_{i,t}(\theta) \hat{A}_i, \ \text{clip}(\bar{r}_{i,t}(\theta), 1-\epsilon, 1+\epsilon) \hat{A}_i \right]$ \\



\bottomrule
\end{tabularx}
\end{sidewaystable*}


\newcommand{\cmark}{\ding{51}}   
\newcommand{\xmark}{\ding{55}}   
\newcommand{\wmark}{$\triangle$} 

\section{Comparison with Existing Surveys}
\label{sec:appendix_comparison}

To better position our contributions and address related literature, we provide a detailed comparison between our work and existing related surveys in Table \ref{tab:survey_comparison}. 

\begin{table*}[tbp]
\centering
\scriptsize 
\renewcommand{\arraystretch}{1.3} 
\tabcolsep=0.1cm 
\caption{Comparison of our survey with existing related surveys on reasoning and post-training.}
\label{tab:survey_comparison}
\begin{tabularx}{\textwidth}{@{}p{3.5cm} >{\raggedright\arraybackslash}X >{\raggedright\arraybackslash}X >{\raggedright\arraybackslash}X >{\raggedright\arraybackslash}X c >{\raggedright\arraybackslash}X @{}}
\toprule
\textbf{Survey} & \textbf{Primary Focus} & \textbf{SFT Focus} & \textbf{Algorithm-Centric} & \textbf{Analytical Findings} & \textbf{Time Coverage} & \textbf{Key Distinction from Our Work} \\
\midrule
\textbf{Thinking Hats} \citep{chen2025putting} & CoT fine-tuning (SFT+RFT) & \wmark~Partial (SFT+RFT together) & \xmark~No & \wmark~Limited (cognitive perspective) & Through 2024 & Different framework: cognitive psychology vs.\ optimization dynamics \\

\textbf{100 Days After R1} \cite{zhang2025100} & R1 replication studies & \wmark~Partial (distillation focus) & \xmark~No & \wmark~Minimal (field report) & Jan--Apr 2025 & Different scope: data-centric scope vs.\ systematic methodology \\

\textbf{Towards Reasoning Era} \cite{chen2025towards} & Long CoT reasoning & \xmark~Minimal (auxiliary to RL) & \xmark~No (RL-centric) & \wmark~Limited (phenomena description) & 2024--early 2025 & Different paradigm: RL-centric vs.\ SFT-centric \\

\textbf{SFT vs RL} \cite{jiang2026supervised} & Post-training paradigms comparison & \wmark~Partial & \wmark~Light (high-level comparison) & \xmark~No & 2022--2025 & Different specificity: general post-training vs.\ reasoning-specific \\

\textbf{LLM Post-Training} \cite{kumar2026llm} & Post-training pipeline (FT, RL, test-time) & \wmark~Partial (one of many) & \wmark~Minimal (pipeline view) & \xmark~No & 2024--2025 & Different granularity: pipeline overview vs.\ SFT deep-dive \\

\textbf{From System 1 to System 2} \cite{li2025system} & Reasoning LLM development & \xmark~Minimal & \xmark~No & \wmark~Limited (benchmark focus) & 2024--mid 2025 & Different focus: broad reasoning LLM survey vs.\ SFT methodology \\

\textbf{Foundation Models Reasoning} \cite{sun2025survey} & Multimodal reasoning paradigms & \xmark~Minimal & \xmark~No & \wmark~Limited (descriptive taxonomy) & Pre-2024 & Different scope: multimodal reasoning applications vs.\ SFT training methods \\

\textbf{Mathematical Reasoning} \cite{wang2026survey} & Math reasoning domain & \wmark~Partial (not focus) & \xmark~No & \wmark~Minimal & Pre-2025 & Different domain: math-specific vs.\ general reasoning; no SFT methodology focus \\

\textbf{Multi-Step Reasoning} \cite{plaat2025multi} & In-context learning/prompting & \xmark~Minimal & \wmark~Yes (prompt algorithms) & \wmark~Limited & 2021--2024 & Different stage: inference-time prompting vs.\ training-time SFT \\

\textbf{Toward Large Reasoning Models} \cite{xu2025toward} & RL for reasoning (``Reinforced Reasoning'') & \xmark~Minimal (SFT as prerequisite) & \cmark~Yes (RL algorithms) & \wmark~Limited & Early 2024--Feb 2025 & Complementary: RL-centric (they cover RL, we cover SFT) \\

\textbf{Reinforced MLLM} \cite{zhou2025reinforced} & RL-based multimodal reasoning & \xmark~No (RL only) & \wmark~Yes (RL for MLLM) & \xmark~No & 2024--2025 & Different modality: multimodal + RL-only vs.\ LLM SFT focus \\

\textbf{RFT for MLLMs} \cite{sun2025reinforcement} & Reinforcement Fine-Tuning for MLLMs & \xmark~No (RFT only) & \wmark~Yes (RL for MLLM) & \xmark~No & 2024--2025 & Different modality: multimodal + RFT vs.\ LLM SFT \\
\midrule
\textbf{Our Survey} & \textbf{SFT for reasoning} & \cmark\cmark~\textbf{Exclusive focus} & \cmark\cmark~\textbf{Core contribution}  & \cmark\cmark~\textbf{Systematic organization} & \textbf{Jul 2024--May 2026} & \textbf{First dedicated reasoning SFT methodology survey} \\
\bottomrule
\multicolumn{7}{l}{\scriptsize \textbf{Legend:} \xmark~No / Minimal \quad \wmark~Limited / Partial \quad \cmark~Yes / Significant \quad \cmark\cmark~Heavy / Core focus} \\
\end{tabularx}
\end{table*}

\textbf{Summary:} Among 12 related surveys, \textbf{none} provide systematic coverage of SFT methodology for reasoning. Prior surveys either: (a) focus on RL rather than SFT \cite{chen2025towards, xu2025toward, zhou2025reinforced, sun2025reinforcement}, (b) cover SFT only peripherally \cite{chen2025putting, zhang2025100, kumar2026llm, li2025system, sun2025survey, wang2026survey}, (c) focus on inference-time prompting \cite{plaat2025multi}, or (d) are domain-specific \cite{wang2026survey}. Our survey fills this gap as the \textbf{first dedicated treatment of reasoning SFT methodology}.

\section{Quantitative Meta-Analysis on a Shared Baseline}
\label{sec:appendix_quantitative_analysis}

We originally refrained from including a quantitative meta-analysis because \textbf{unconstrained cross-paper comparisons are scientifically unreliable due to vastly different base models, training datasets, and evaluation setups across the literature.} We were concerned that an uncurated aggregation might present misleading conclusions about relative method performance.

However, we agree with the reviewer that an empirical synthesis adds substantial practical value. To provide a rigorous and fair comparison, \textbf{we have now added a quantitative analysis focused strictly on methods sharing the exact same baseline: Qwen2.5-Math-7B fine-tuned on NuminaMath-CoT.} We report their macro-average performance across four shared benchmarks (Minerva Math, OlympiadBench, AIME 2024, and AMC 2023), collecting results from the DFT, RIFT, RASFT, and RANKTUNER papers. Note that RFT* and RIFT* are evaluated as exceptions, as they train on self-generated responses rather than the static NuminaMath-CoT annotations.

\begin{table*}[htbp]
\centering
\caption{Quantitative comparison of methods sharing the Qwen2.5-Math-7B fine-tuned on NuminaMath-CoT.}
\label{tab:quantitative_comparison}
\resizebox{\textwidth}{!}{
\begin{tabular}{lcccccccc}
\toprule
\textbf{Method} & \textbf{GSM8K} & \textbf{MATH} & \textbf{Minerva} & \textbf{Olympiad} & \textbf{AIME 2024} & \textbf{AMC 2023} & \textbf{Avg. over 4} \\
\midrule
No training & 54.8 & 50.3 & 14.4 & 16.4 & 12.1 & 36.9 & 20.0 \\
EAFT & -- & -- & 20.2 & 19.0 & 2.5 & 24.2 & 16.5 \\
ASFT & -- & 65.8 & 24.1 & 29.3 & 9.0 & 43.3 & 26.4 \\
SFT & 71.1 & 60.9 & 21.8 & 32.9 & 9.2 & 43.4 & 26.8 \\
ProFiT & -- & 68.5 & 28.7 & 31.4 & 7.1 & 44.1 & 27.8 \\
DFT & 87.0 & 70.6 & 30.2 & 33.8 & 8.6 & 45.0 & 29.4 \\
RASFT & -- & 70.7 & 30.0 & 33.0 & 8.8 & 45.8 & 29.4 \\
TALR & -- & -- & 35.7 & 32.9 & 6.7 & 43.4 & 29.7 \\
RFT* & 83.3 & 69.8 & 21.3 & 31.3 & 11.2 & 58.8 & 30.7 \\
iw-SFT & -- & 70.3 & 25.7 & 34.5 & 16.5 & 51.1 & 32.0 \\
RANKTUNER & -- & -- & 38.3 & 36.3 & 10.2 & 46.6 & 32.9 \\
RIFT* & 86.3 & 74.7 & 28.9 & 34.1 & 17.1 & 62.2 & 35.6 \\
\bottomrule
\end{tabular}
}
\end{table*}

Three main observations emerge from this synthesis:

\begin{enumerate}
    \item \textbf{Modified SFT objectives do not guarantee gains.} EAFT (16.5) and ASFT (26.4) underperform vanilla SFT (26.8), highlighting that the success of reweighting or regularization heavily depends on how well its inductive bias matches the task.
    
    \item \textbf{Early baselines remain remarkably strong.} Simpler, earlier methods like DFT (29.4) and iw-SFT (32.0) outperform several recently proposed, more complex variants. Chronological novelty does not strictly translate to better performance.
    
    \item \textbf{Self-generated CoTs outperform external trajectories.} RFT* and RIFT*, which train on self-generated responses, achieve the highest averages (30.7 and 35.6). This suggests that self-generated trajectories better match the model's native policy distribution and are therefore easier to internalize than externally provided CoTs.
\end{enumerate}

We note that these observations are derived strictly from this specific model and dataset configuration, serving as an \textbf{empirical reference rather than universal conclusions}.

\end{document}